\documentclass[journal]{IEEEtran}

\ifCLASSINFOpdf

\else

\fi

\usepackage{booktabs}
\usepackage{multirow}
\usepackage{array}

\usepackage{graphicx}  
\usepackage{float}     
\usepackage{caption}   
\usepackage{xcolor}

\usepackage{subcaption}  
\usepackage{stfloats}   
\usepackage{hyperref}
\hypersetup{citebordercolor=green}
\hypersetup{linkbordercolor=red}
\hypersetup{urlbordercolor=blue}
\usepackage{amsmath}
\usepackage{amssymb}
\usepackage{dsfont}

\begin{document}

\newcommand\hai[1]{\textbf{\textcolor{red}{Zhu: #1}}}

\title{Vision-Language Navigation for Aerial Robots: Towards the Era of Large Language Models}

\author{Xingyu Xia$^1$, Lekai Zhou$^2$, Yujie Tang$^{2,*}$, Xiaozhou Zhu$^1$, Hai Zhu$^{1,*}$, and Wen Yao$^{1,*}$
\thanks{This work was partially supported by the National Natural Science Foundation of China (62303486, 92371206).}
\thanks{$^1$Xingyu Xia, Xiaozhou Zhu, Hai Zhu, and Wen Yao are with the Defense Innovation Institute, Chinese Academy of Military Sciences, Beijing 10071, China, and are also with the Intelligent Game and Decision Laboratory, Beijing 100071, China.}
\thanks{$^2$Lekai Zhou, and Yujie Tang are with the Aerospace Information Research Institute, Chinese Academy of Sciences, Beijing 100094, China.}
\thanks{$^*$Corresponding authors. ({\tt\small zhuhai11@alumni.nudt.edu.cn; tangyj@aircas.ac.cn; yaowen@nudt.edu.cn})} 
}


\maketitle

\begin{abstract}
Aerial vision-and-language navigation (Aerial VLN) aims to enable unmanned aerial vehicles (UAVs) to interpret natural language instructions and autonomously navigate complex three-dimensional environments by grounding language in visual perception. Unlike ground-based VLN, the aerial setting introduces multiple qualitatively distinct challenges, including a six-degree-of-freedom continuous action space, severe viewpoint variation driven by altitude and orientation changes, city-scale navigation with lengthy and structurally complex instructions, and onboard computational constraints imposed by lightweight platforms. This survey provides a critical and analytical review of the Aerial VLN field, with particular attention to the recent integration of large language models (LLMs) and vision-language models (VLMs).  We first formally introduce the Aerial VLN problem and define two interaction paradigms: single-instruction (AVIN) and dialog-based (AVDN), as foundational axes.  We then organize the body of Aerial VLN methods into a taxonomy of five architectural categories: sequence-to-sequence and attention-based methods, end-to-end LLM/VLM methods, hierarchical methods, multi-agent methods, and dialog-based navigation methods. For each category, we systematically analyze design rationales, technical trade-offs, and reported performance. We critically assess the evaluation infrastructure for Aerial VLN, including datasets, simulation platforms, and metrics, and identify their gaps in scale, environmental diversity, real-world grounding, and metric coverage.  We consolidate cross-method comparisons on shared benchmarks and analyze key architectural trade-offs, including discrete versus continuous actions, end-to-end versus hierarchical designs, and the simulation-to-reality gap.  Finally, we synthesize seven concrete open problems: long-horizon instruction grounding, viewpoint robustness, scalable spatial representation, continuous 6-DoF action execution, onboard deployment, benchmark standardization, and multi-UAV swarm navigation, with specific research directions grounded in the evidence presented throughout the survey.
\end{abstract}

\begin{IEEEkeywords}
Vision-language navigation, UAV, Large language models, Vision language models, Vision foundation models, Autonomous navigation.
\end{IEEEkeywords}

\IEEEpeerreviewmaketitle

\section{Introduction}

\subsection{Motivation}
Unmanned aerial vehicles (UAVs) have become indispensable platforms across a growing range of domains---from intelligent transportation and logistics~\cite{LiIntelligentTransportation, BettieliverySystems} to precision agriculture~\cite{LiuPrecisionAgriculture}, source seeking~\cite{huang2025online} and infrastructure inspection~\cite{zhu2021online}, driven by their spatial mobility, flexible viewpoints, and rapid deployability~\cite{zhuEdgeComputingPowers2023, huangSmalldroneRevolutionComing2025}. As these applications scale in scope and complexity, so too does the demand for UAV systems that can operate with greater autonomy, adapt to unstructured environments, and interact naturally with human operators~\cite{tianUAVsMeetLLMs2025}.

\begin{figure}[t]  
    \centering
    \includegraphics[width=0.9\linewidth]{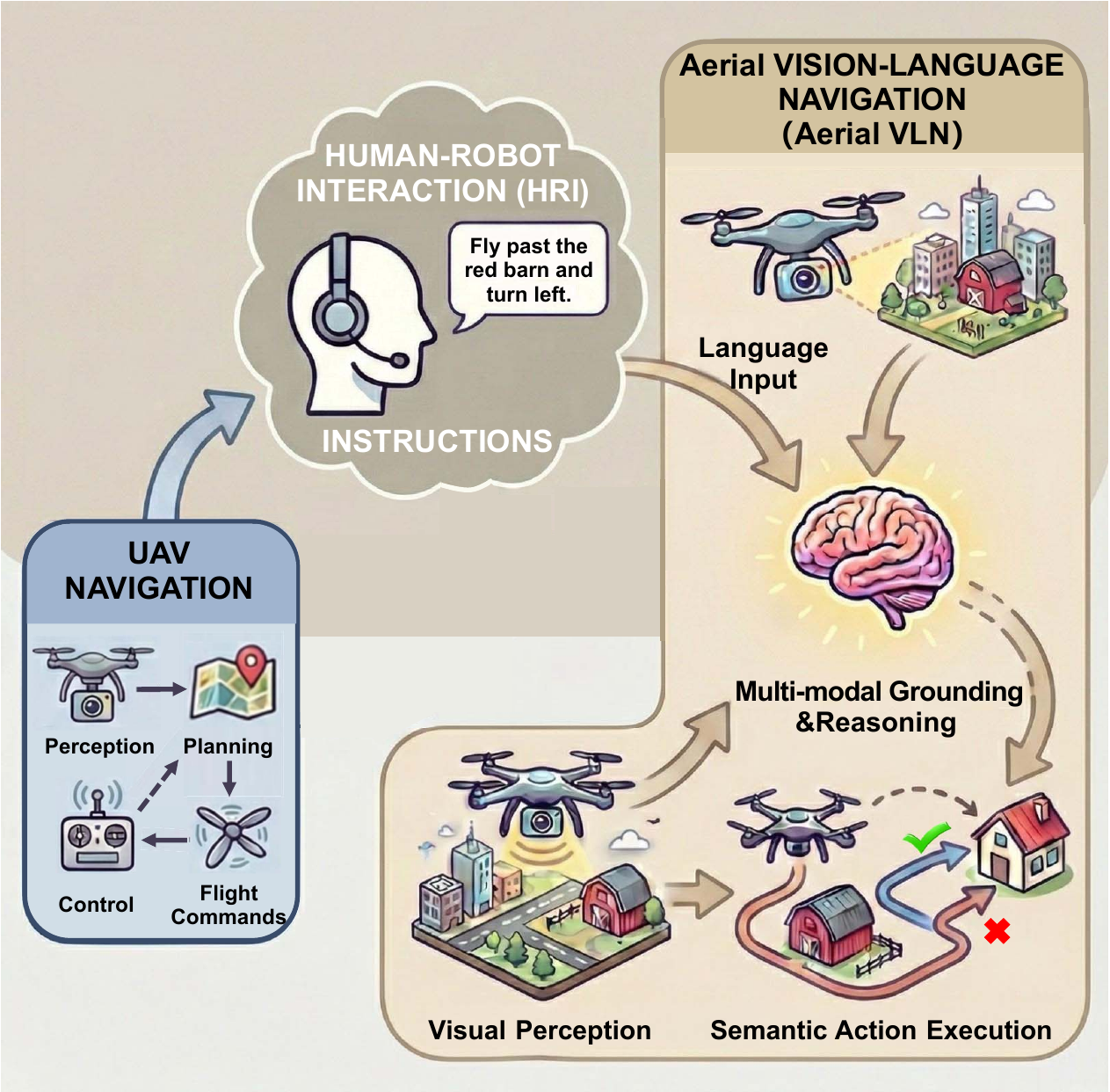}  
    \caption{The evolution from UAV Navigation to Aerial VLN. Classical UAV navigation relies on the perception-planning-control loop to execute flight commands. Triggered by human-robot interaction (HRI), specifically natural-language instructions, Aerial VLN aligns visual perception with instructions, enabling UAVs to perform semantic action execution rather than solely following planned trajectories.}
    \label{fig:Navigation_preliminary}
\end{figure}

A central bottleneck in achieving this autonomy is the interface between human intent and robot action, known as HRI. As illustrated in Fig. \ref{fig:Navigation_preliminary}, conventional UAV navigation systems follow a modular perception--planning--control pipeline to execute predefined flight commands~\cite{NahavandiComprehensiveReviewAutonomous2025, Ren2025Survey}, or learning-based methods for module integration and vision-action mapping~\cite{loquercioLearningHighspeedFlight2021, heExplainableDeepReinforcement2021a}. While effective for structured tasks with explicit waypoints, this architecture lacks the capacity for high-level task reasoning: specifically, translating natural language instructions into a grounded sequence of actions informed by real-time visual perception. And this is the core problem of aerial vision-and-language navigation (Aerial VLN), reasoning over instructions, parsing the scene, and executing actions~\cite{liuAerialVLNVisionandlanguageNavigation2023, fanAerialVisionanddialogNavigation2023a}.

Aerial VLN is not simply the application of ground-based VLN methods to a flying platform. Structural properties of the aerial domain like state space expansion, viewpoint variation and larger spatial scale, create qualitatively distinct challenges compared with ground VLN.
The practical significance of solving Aerial VLN is substantial. Language-guided UAV patrol in smart city management~\cite{zheng2026onfly, zhang2026apex}, last-mile delivery UAVs in logistics~\cite{zhangLogisticsVLNVisionlanguageNavigation2025}, fire fighting UAVs~\cite{pingMultimodalLargeLanguage2025} and so on are potential applications in Aerial VLN. Across these scenarios, Aerial VLN serves as the enabling capability that transforms UAVs from remotely operated tools into autonomous partners capable of understanding and acting on human intent.

Recent advances in large language models (LLMs), including GPT-4~\cite{brownLanguageModelsAre2020}, DeepSeek~\cite{guoDeepSeekR1IncentivizesReasoning2025}, and Qwen~\cite{baiQwenTechnicalReport2023}, and in vision foundation models (VFMs) such as CLIP~\cite{radfordLearningTransferableVisual2021}, SAM~\cite{kirillovSegmentAnything2023}, and Grounding DINO~\cite{liuGroundingDINOMarrying2025} have catalyzed a paradigm shift in how Aerial VLN systems are designed. Emerging indoor and ground VLN works~\cite{Zhao2026NavGeminiAM, Huang2025UNeMoCV, schumannVELMAVerbalizationEmbodiment2024, shahLMnavRoboticNavigation2023} have also demonstrated the significant potential of integrating LLMs into VLN. Where classic methods relied on task-specific encoders trained from scratch on limited aerial datasets, LLM-centric approaches leverage the semantic reasoning, world knowledge, and zero-shot generalization capabilities of large pre-trained models to serve as the cognitive core of the navigation system~\cite{tianUAVsMeetLLMs2025,yaoAeroVersereviewComprehensiveSurvey2025}. This integration has given rise to new architectural paradigms: end-to-end methods that directly map instructions and perception to actions~\cite{caiFlightGPTGeneralizableInterpretable2025,liuNavAgentMultiscaleUrban2024}, hierarchical methods that pair LLM-based planners with traditional flight controllers~\cite{liSkyVLNVisionandlanguageNavigation2025, gaoOpenFlyComprehensivePlatform2025a, wangHiAirStarGuide2025}, and multi-agent methods that distribute reasoning across collaborating LLM agents~\cite{sautenkovUAVCodeAgentsScalableUAV2025,zhangMMCNavMLLMempoweredMultiagent2025}. These developments have expanded the frontier of what is achievable in Aerial VLN, but introduced new challenges in computational efficiency, sim-to-real transfer, and the coupling of semantic reasoning with robust flight control.

Despite this rapid progress, the research landscape of Aerial VLN remains fragmented. Methods are evaluated on disparate benchmarks with incompatible metrics, making cross-method comparison difficult. Several existing VLN surveys~\cite{guVisionandlanguageNavigationSurvey2022, wuVisionlanguageNavigationSurvey2022, zhangVisionandlanguageNavigationToday2024} provide valuable coverage of ground-based tasks but either omit aerial platforms entirely or treat them as a peripheral extension. The concurrent AeroVerse-Review~\cite{yaoAeroVersereviewComprehensiveSurvey2025} offers broad coverage of Aerial VLN but introduces a degree of ambiguity in method classification, and doesn't provide the quantitative cross-method comparison. Tian et al.~\cite{tianUAVsMeetLLMs2025} survey the broader integration of LLMs with UAVs but don't focus specifically on the VLN problem or its unique technical challenges. These gaps motivate our survey, which aims to provide not merely a catalog of existing works but a critical synthesis---comparing methods on shared benchmarks where possible, evaluating the adequacy of current datasets and simulation platforms, and identifying the sim-to-real open problems in Aerial VLN.

\subsection{Scope}
\label{sec:scope}

To ensure transparent and reproducible coverage, we define the boundaries and selection criteria of this survey as follows.


For literature sources, we draw on both peer-reviewed publications and pre-prints.  Peer-reviewed sources include journals and conference proceedings from IEEE, ACM, AAAI, NeurIPS, ICLR, CVPR, ICCV, ECCV, and ICRA/IROS, covering the intersecting fields of vision-language navigation, large language models, and UAV systems.  Given the rapid pace of development in LLM-integrated aerial navigation, we also include pre-prints from arXiv when they introduce methods, datasets, or benchmarks that have gained visible community adoption (e.g., cited by subsequent peer-reviewed works or accompanied by public code releases).  Pre-print status is not treated as equivalent to peer review; where pre-prints and peer-reviewed works cover similar contributions, we prioritize the latter.


Regarding the literature time period, the core time frame spans from 2018 when the first Aerial VLN datasets and methods appeared~\cite{misraMappingInstructionsActions2018, blukisMappingNavigationInstructions}, to 2026.  Within this window, the primary focus is on work published from 2023 onward, coinciding with the integration of LLMs into Aerial VLN and the resulting paradigm shift from feature-matching-based navigation to reasoning-based navigation.  Earlier foundational works in indoor VLN, outdoor VLN, and UAV control are cited for context where necessary but are not reviewed in depth, as comprehensive surveys of these areas already exist~\cite{NahavandiComprehensiveReviewAutonomous2025, guVisionandlanguageNavigationSurvey2022, wuVisionlanguageNavigationSurvey2022, zhangVisionandlanguageNavigationToday2024}.


The core focus of this survey is Aerial VLN: methods, datasets, platforms, and metrics that address the problem of UAV navigation guided by natural language instructions combined with visual perception.  Three boundary decisions shape the coverage:

\begin{itemize}
    \item \textit{Ground-based VLN} is discussed in Section~\ref{sec:context} for comparative positioning but is not reviewed method-by-method; readers are directed to dedicated surveys~\cite{guVisionandlanguageNavigationSurvey2022, wuVisionlanguageNavigationSurvey2022, zhangVisionandlanguageNavigationToday2024}.
    \item \textit{UAV navigation without language grounding} (e.g., waypoint-based path planning, pure reinforcement learning for obstacle avoidance) falls outside the scope unless a method directly contributes to the Aerial VLN pipeline as a low-level controller or baseline.
    \item \textit{General-purpose LLM and VLM architectures} (e.g., GPT-4, LLaVA) are referenced as building blocks but are not independently reviewed; the focus is on how they are adapted and deployed within Aerial VLN systems.
\end{itemize}

For datasets and benchmarks, we emphasize resources that are publicly available and have been adopted in multiple published works.  Domain-specific datasets with potential but undemonstrated applicability to Aerial VLN are cataloged separately as underexploited resources (Section~\ref{sec:datasets}).

\subsection{Contributions and Organization}
The main contributions of this paper are as follows:

\begin{enumerate}

\item \textbf{A unified architectural taxonomy of Aerial VLN methods.}
We organize the Aerial VLN research into five categories defined by architectural principle: sequence-to-sequence and attention-based methods, end-to-end LLM/VLM methods, hierarchical methods, multi-agent methods, and dialog-based navigation methods. Within each category, we trace the evolution from early designs to current LLM-integrated approaches, making the continuity and progression of ideas visible.  

\item \textbf{A critical assessment of evaluation infrastructure.}
We provide a standalone, structured analysis of the datasets, simulation platforms, and evaluation metrics that underpin Aerial VLN research.  Beyond cataloging what exists, we evaluate the adequacy of the current infrastructure: we identify gaps in dataset scale and diversity, compare simulators along dimensions specific to aerial navigation fidelity, and critically examine the prevailing metrics, including the adoption in current Aerial VLN works.

\item \textbf{A quantitative and qualitative comparative analysis.}
We consolidate reported results from multiple methods on shared benchmarks into performance comparison tables, and provide cross-cutting analysis of key architectural trade-offs. We further discuss the sim-to-real gap by cataloging which methods have been validated on physical UAV platforms. This comparative synthesis constitutes the analytical core of the paper.

\item \textbf{A thematically organized roadmap of open problems.}
We identify seven concrete open problems: spanning long-horizon instruction grounding, viewpoint robustness, scalable spatial representation, continuous 6-DoF action execution, onboard deployment efficiency, benchmark standardization, and multi-UAV swarm navigation. And for each open problem, we review what current approaches have attempted, explain their limitations, and propose specific research directions.

\end{enumerate}

\begin{figure*}[t!]  
    \centering
    \includegraphics[width=1\linewidth]{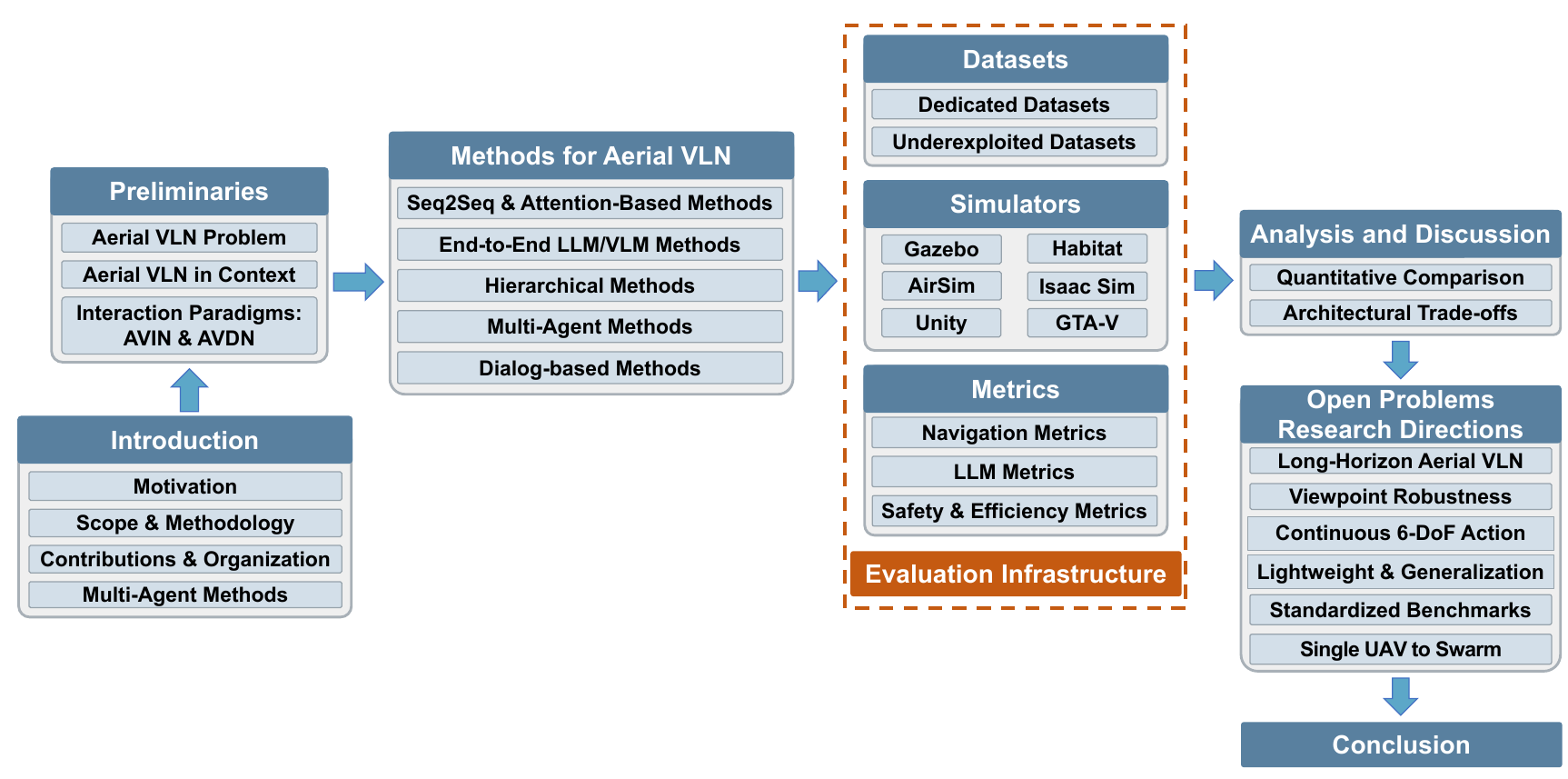}  
    \caption{Structure and organization of the survey paper.}
    \label{fig:Article_structure}
\end{figure*}

The remainder of this paper is organized as follows: Section~\ref{sec:preliminaries} formulates the Aerial VLN problem, delineates features relative to indoor/outdoor ground VLN, and outlines the two primary interaction paradigms. Section~\ref{sec:methods} reviews Aerial VLN methods under the unified architectural taxonomy.  Section~\ref{sec:infrastructure} critically assesses the evaluation infrastructure, including datasets, simulation platforms, and metrics. Section~\ref{sec:analysis} presents qualitative/quantitative comparison analysis. Section~\ref{sec:openproblems} synthesizes open problems and future research directions. Section~\ref{sec:conclusions} concludes the paper. Fig. \ref{fig:Article_structure} illustrates the structure and organization of this survey paper. 

\section{Preliminaries}\label{sec:preliminaries}

\begin{figure*}[t]  
    \centering
    \includegraphics[width=1\linewidth]{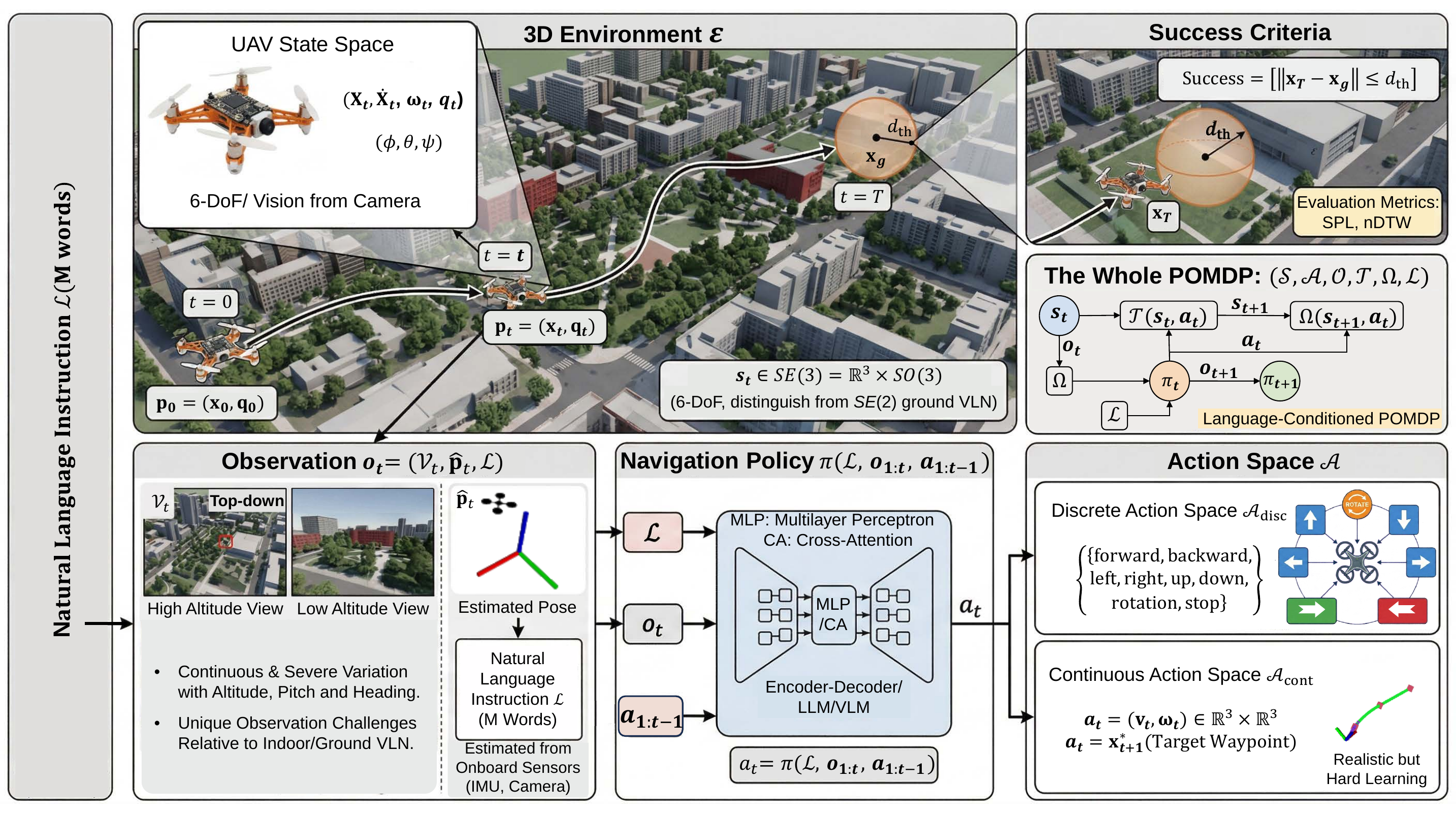}  
    \caption{The problem definition of Aerial VLN.}
    \label{fig:Problem_definition}
\end{figure*}

\subsection{Aerial VLN Problem}\label{sec:formulation}
The Aerial VLN problem is typically formulated as a language-conditioned sequential decision-making problem in partially observable 3D space. The formulation captures the essential elements that any Aerial VLN method must address: state representation, perception under partial observability, action generation, and task success evaluation, while remaining general enough to encompass both discrete and continuous instantiations. Fig. \ref{fig:Problem_definition} gives an illustration of the Aerial VLN problem, which is formally defined in the following. 

\subsubsection{Task Definition}
An Aerial VLN episode proceeds as follows.  At the start of an episode, the UAV is initialized at a pose $\mathbf{p}_0 = (\mathbf{x}_0, \mathbf{q}_0)$ in a 3D environment $\mathcal{E}$, where $\mathbf{x}_0 \in \mathbb{R}^3$ denotes position and $\mathbf{q}_0 \in SO(3)$ denotes orientation.  A natural language instruction $\mathcal{L} = (w_1, w_2, \ldots, w_M)$, consisting of $M$ words, specifies the intended navigation task, which is either as a step-by-step route description or as a goal-oriented target specification. The UAV must interpret $\mathcal{L}$ in conjunction with its visual observations to navigate through $\mathcal{E}$ and reach a goal location $\mathbf{x}_g$ that satisfies the instruction. At each discrete time step $t = 1, 2, \ldots, T$, the UAV selects an action $a_t$ based on the instruction $\mathcal{L}$, the history of observations $o_{1:t}$, and its state $s_t$. The episode terminates when the UAV issues a \texttt{STOP} action or when a maximum step budget $T_{\max}$ is reached.

\subsubsection{State Space}

The full state of the system at time $t$ is defined as:
\begin{equation}
    s_t = (\mathbf{x}_t, \dot{\mathbf{x}}_t, \boldsymbol{\omega}_t, \mathbf{q}_t, \mathcal{E})
    \label{eq:state}
\end{equation}
where $\mathbf{x}_t \in \mathbb{R}^3$, $\dot{\mathbf{x}}_t \in \mathbb{R}^3$ and $\boldsymbol{\omega}_t \in \mathbb{R}^3$ are the UAV's position, linear velocity and angular velocity, $\mathbf{q}_t \in SO(3)$, or parameterized by Euler angles $(\phi, \theta, \psi)$ is its orientation, and $\mathcal{E}$ represents the full environment state including the geometry, semantics, and dynamics of all objects. The UAV's pose thus evolves in 6-DoF configuration space $SE(3) = \mathbb{R}^3 \times SO(3)$, a fundamental distinction from ground VLN agents that operate in $SE(2) = \mathbb{R}^2 \times S^1$.

In practice, the UAV cannot observe $s_t$ in its entirety. The environment state $\mathcal{E}$ is typically unknown a priori, the UAV state ($\mathbf{x}_t$, $\dot{\mathbf{x}}_t$, $\boldsymbol{\omega}_t$, $\mathbf{q}_t$) is real-time available through onboard estimation, and visual perception is limited by the sensor's field of view. This partial observability makes the Aerial VLN problem naturally suited to a POMDP formulation~\cite{chenGRADNAVVisionlanguageModel2026}.

\subsubsection{Observation Space}

At each time step, the UAV receives an observation $o_t$ that provides a partial and noisy window into the true state $s_t$. We decompose the observation into three components:
\begin{equation}
    o_t = (\mathcal{V}_t, \hat{\mathbf{p}}_t, \mathcal{L})
    \label{eq:observation}
\end{equation}
where $\mathcal{V}_t$ denotes the visual observation, $\hat{\mathbf{p}}_t$ denotes the estimated UAV pose, and $\mathcal{L}$ is the language instruction (constant throughout the episode in the single-instruction paradigm, or augmented by new dialog turns $d_t$ in the dialog paradigm, see Section~\ref{sec:interaction}).

The visual observation $\mathcal{V}_t$ depends on the sensor configuration. $\mathcal{V}_t$ may consist of a single egocentric RGB image $I_t \in \mathbb{R}^{H \times W \times 3}$, a set of multi-view images $\{I_t^{(k)}\}_{k=1}^K$~\cite{zhaoAerialVisionandlanguageNavigation2025}, RGB-D data augmented with depth $D_t \in \mathbb{R}^{H \times W}$, or point cloud data from LiDAR. A critical characteristic of Aerial VLN is that $\mathcal{V}_t$ undergoes continuous and severe variation as the UAV changes altitude, pitch, and heading during flight. The same physical landmark can produce drastically different image appearances across consecutive time steps, which is a challenge that has no close analogue in indoor or street-level ground VLN, where the camera height and viewing angle remain approximately constant.

\subsubsection{Action Space}

The action space $\mathcal{A}$ defines the set of commands the UAV can execute at each time step. Existing Aerial VLN methods instantiate $\mathcal{A}$ in two fundamentally different ways:

\paragraph{Discrete action space.}  The majority of current methods~\cite{liuAerialVLNVisionandlanguageNavigation2023, caiFlightGPTGeneralizableInterpretable2025, liuNavAgentMultiscaleUrban2024, gaoAerialVisionandlanguageNavigation2024} define a finite set of primitive actions:
\begin{equation}
    \begin{split}
    \mathcal{A}_{\text{disc}} = \{\texttt{forward}, \texttt{backward}, \texttt{left}, \texttt{right}, \\
    \texttt{up}, \texttt{down},  \texttt{rotation}, \texttt{stop}\}
    \label{eq:discrete_action}
    \end{split}
\end{equation}
Each action displaces the UAV by a fixed distance or rotates it by a fixed angle.  While this discretization simplifies the learning problem and aligns naturally with the token-generation paradigm of LLMs, it imposes an artificial constraint on the UAV's motion: real flight is inherently continuous, and coarse discretization can lead to jerky trajectories, imprecise goal reaching, and unrealistic navigation behavior~\cite{wangRealisticUAVVisionlanguage2024}.

\paragraph{Continuous action space.}  A smaller but growing set of methods~\cite{gaoOpenFlyComprehensivePlatform2025a, blukisMappingNavigationInstructions, chenGRADNAVVisionlanguageModel2026} operate in a continuous action space:
\begin{equation}
    a_t = (\mathbf{v}_t, \boldsymbol{\omega}_t) \in \mathbb{R}^3 \times \mathbb{R}^3
    \label{eq:continuous_action}
\end{equation}
where $\mathbf{v}_t$ is a velocity command and $\boldsymbol{\omega}_t$ is an angular rate command, or equivalently a target waypoint $\mathbf{x}_{t+1}^* \in \mathbb{R}^3$ to be tracked by a low-level flight controller.  Continuous action spaces are more faithful to real UAV dynamics but dramatically increase the difficulty of the learning problem, particularly when actions must be inferred from language instructions rather than from dense reward signals.

The tension between discrete and continuous action spaces is a defining characteristic of Aerial VLN methods, and also a recurrent theme throughout this survey. We revisit this trade-off in detail in Sections~\ref{sec:methods} and~\ref{sec:analysis}.

\subsubsection{Navigation Policy}

The goal of an Aerial VLN agent is to learn a navigation policy $\pi$ that maps the instruction and the history of observations to an action at each time step:
\begin{equation}
    a_t = \pi(\mathcal{L}, o_{1:t}, a_{1:t-1})
    \label{eq:policy}
\end{equation}

In classic methods, $\pi$ is typically parameterized as an encoder--decoder architecture: the instruction $\mathcal{L}$ and visual observations $\mathcal{V}_{1:t}$ are encoded into feature vectors via separate language and vision encoders, fused through attention mechanisms, and decoded into an action~\cite{liuAerialVLNVisionandlanguageNavigation2023, blukisMappingNavigationInstructions}.  In LLM-centric methods, $\pi$ may be realized as a prompted LLM that receives states and outputs actions~\cite{caiFlightGPTGeneralizableInterpretable2025, gaoAerialVisionandlanguageNavigation2024}, as a VLM that processes raw images alongside instructions~\cite{liuNavAgentMultiscaleUrban2024, gaoOpenFlyComprehensivePlatform2025a}, or as a hierarchical system in which LLMs generate high-level subgoals that are executed by a separate low-level controller~\cite{liSkyVLNVisionandlanguageNavigation2025, wangHiAirStarGuide2025,zhangCityNavAgentAerialVisionandlanguage2025}.  Section~\ref{sec:methods} organizes these diverse instantiations into a unified architectural taxonomy.

\subsubsection{Success Criteria}

An episode is judged successful if the UAV's final position $\mathbf{x}_T$ lies within a threshold distance $d_{\text{th}}$ of the goal location $\mathbf{x}_g$. We introduce the indicator function $\mathds{1}[\cdot]$ which equals 1 if the condition is true and 0 otherwise:
\begin{equation}
    \text{Success} = \mathds{1} \left[\|\mathbf{x}_T - \mathbf{x}_g\|_2 \leq d_{\text{th}}\right]
    \label{eq:success}
\end{equation}
The threshold $d_{\text{th}}$ varies across benchmarks: the AerialVLN dataset~\cite{liuAerialVLNVisionandlanguageNavigation2023} uses $d_{\text{th}} = 20$ m to reflect the large spatial scale of city-level navigation, while CityNav~\cite{leeCityNavLargescaleDataset2025} adopts a tighter $d_{\text{th}}$ suited to scenarios represented by point cloud. Beyond this binary success, the quality of a navigation episode is further characterized by path efficiency (SPL~\cite{anderson2018vision}), trajectory fidelity (nDTW), and additional metrics discussed in Section~\ref{sec:metrics}.

\subsubsection{POMDP Formulation}

The formulation above can be compactly expressed as a language-conditioned POMDP defined by the tuple $(\mathcal{S}, \mathcal{A}, \mathcal{O}, \mathcal{T}, \Omega, \mathcal{L})$, where $\mathcal{S}$ is the state space (Eq.~(\ref{eq:state})), $\mathcal{A}$ is the action space (Eq.~(\ref{eq:discrete_action}) or~(\ref{eq:continuous_action})), $\mathcal{O}$ is the observation space (Eq.~(\ref{eq:observation})), $\mathcal{T}: \mathcal{S} \times \mathcal{A} \rightarrow \Delta(\mathcal{S})$ is the state transition function governed by the UAV's dynamics and the environment, $\Omega: \mathcal{S} \times \mathcal{A} \rightarrow \Delta(\mathcal{O})$ is the observation function determined by the sensor model, and $\mathcal{L}$ is the conditioning instruction. Unlike a standard POMDP, no explicit scalar reward function $R$ is defined. Instead, the agent optimizes for instruction-following behavior, which is typically supervised via imitation learning (IL) on expert trajectories or shaped through task-specific reward signals during reinforcement learning (RL).

This formulation highlights properties that distinguish Aerial VLN from ground-based VLN as a decision-making problem. In state space, $\mathcal{S}$ operates over $SE(3)$ rather than $SE(2)$, introducing tightly coupled spatial dynamics. In observation space, $\Omega$ produces highly variable outputs due to altitude-dependent viewpoint shifts, making cross-modal alignment between $\mathcal{V}_t$ and $\mathcal{L}$ substantially more difficult. Additionally, the spatial scale of aerial environment and the corresponding length of instructions create long-horizon dependency structures.

\begin{figure*}[t]  
    \centering
    \includegraphics[width=0.9\linewidth]{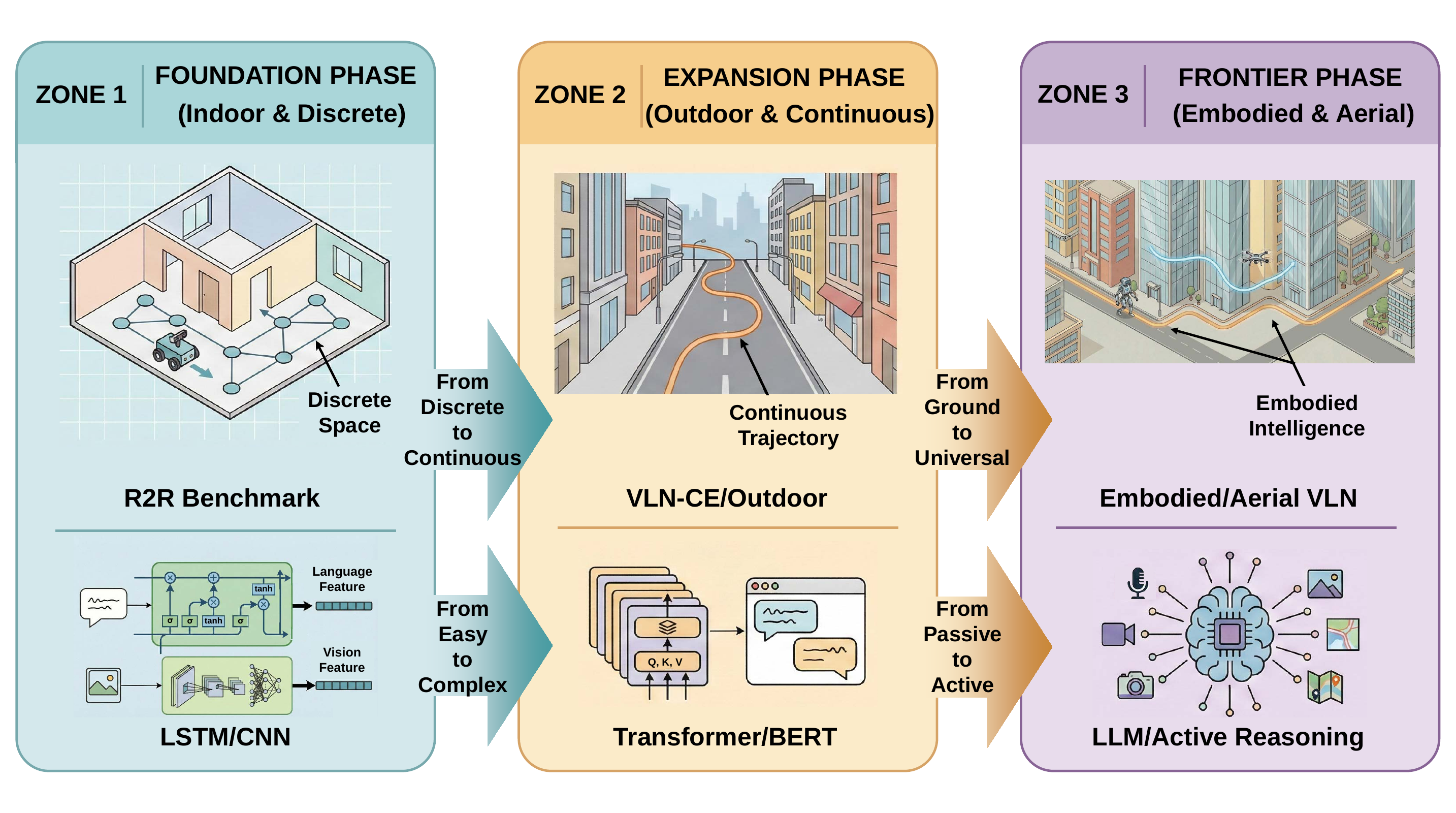}  
    \caption{The evolution of VLN. It progresses from the foundation phase, characterized by discrete indoor navigation and LSTM/CNN architectures, to the expansion phase involving continuous outdoor trajectories processed by Transformer. For future frontier phase, LLM-driven active reasoning empowers embodied and aerial agents to achieve universal adaptability in complex environments.}
    \label{fig:VLN_preliminary}
\end{figure*}

\begin{table*}[t]
\centering
\caption{Comparative characteristics of VLN across indoor, outdoor ground, and aerial settings.  Representative benchmarks are listed for each setting.  Values are approximate and based on dataset statistics reported in the cited works.}
\label{tab:vln_comparison}
\setlength{\tabcolsep}{2pt}
\small
\begin{tabular}{l c c c}
\toprule
\textbf{Dimension} & \textbf{Indoor VLN} & \textbf{Outdoor Ground VLN} & \textbf{Aerial VLN} \\
\midrule
\textit{Representative benchmarks} & R2R~\cite{anderson2018vision}, VLN-CE~\cite{krantzNavgraphVisionandlanguageNavigation2020} & Touchdown~\cite{chenTOUCHDOWNNaturalLanguage2019}, StreetLearn~\cite{mirowskiStreetLearnEnvironmentDataset2019} & \begin{tabular}{@{}c@{}}AerialVLN~\cite{liuAerialVLNVisionandlanguageNavigation2023}, CityNav~\cite{leeCityNavLargescaleDataset2025} \\ OpenFly~\cite{gaoOpenFlyComprehensivePlatform2025a}, AVDN~\cite{fanAerialVisionanddialogNavigation2023a}\end{tabular} \\
\addlinespace
\textit{Configuration space} & $SE(2)$: 3-DoF & $SE(2)$: 3-DoF & $SE(3)$: 6-DoF \\
\textit{Action space} & Discrete node-based movement or & Discrete panoramic turns or & Discrete 6-direction primitives or \\
                       & low-level continuous ($v$, $\omega$) & street-level steps & continuous 6-DoF ($v$, $\omega$) \\
\addlinespace
\textit{Typical trajectory length} & 5--15\,m & 100--1000\,m & 100--500+\,m \\
\textit{Avg.\ instruction length} & $\sim$29 words (R2R) & $\sim$80 words (Touchdown) & $\sim$80--180 words (AerialVLN, OpenFly) \\
\textit{Success threshold $d_{\text{th}}$} & 3\,m & 10--40\,m & 15--20\,m \\
\addlinespace
\textit{Viewpoint dynamics} & Nearly fixed height, & Fixed height, & Continuous altitude, pitch, \\
                             & horizontal rotation only & horizontal panning & and heading variation \\
\textit{Environment type} & Scanned indoor rooms & Street-level panoramas & City-scale 3D reconstructions \\
                           & (Matterport3D, Gibson) & (Google Street View) & or game-engine worlds (UE, GTA-V) \\
\addlinespace
\textit{Computational deployment} & Offboard (GPU server) & Offboard (GPU server) & Offboard or onboard (edge GPU) \\
\bottomrule
\end{tabular}
\end{table*}

\subsection{Aerial VLN in Context}
\label{sec:context}

VLN is a representative embodied intelligence task~\cite{DuanSurveyEmbodiedAI}, requires agents to achieve cross-modal alignment and reasoning among semantics, vision, and navigation~\cite{ShuangLinSIMARecentAdvancesVisionandlanguage2023, dosovitskiyImageWorth16x162021}. VLN has progressed through three overlapping phases: indoor discrete navigation, outdoor continuous navigation, and aerial navigation, as illustrated in Fig. \ref{fig:VLN_preliminary}.  Rather than recounting this history in full, for which comprehensive surveys already exist~\cite{guVisionandlanguageNavigationSurvey2022, wuVisionlanguageNavigationSurvey2022, zhangVisionandlanguageNavigationToday2024, parkVisualLanguageNavigation2023}, we focus here on the structural differences that make Aerial VLN a qualitatively distinct problem.  Table~\ref{tab:vln_comparison} summarizes these differences across five dimensions, which are analyzed in the following specific aspects.

\subsubsection{Degrees of Freedom and Action Complexity}

Indoor VLN, typified by the Room-to-Room (R2R) benchmark~\cite{anderson2018vision}, or  its methodological derivatives~\cite{jainStayPathInstruction2019, hongLanguageVisualEntity2020}, were formulated on discrete navigation graphs in which the agent teleports between pre-defined viewpoints.  Even in continuous-environment extensions such as VLN-CE~\cite{krantzNavgraphVisionandlanguageNavigation2020}, the agent moves on a horizontal plane with 3-DoF (position $x$, $y$ and heading $\theta$). Outdoor ground VLN datasets~\cite{mirowskiStreetLearnEnvironmentDataset2019, chenTOUCHDOWNNaturalLanguage2019, hermannLearningFollowDirections2020} similarly constrain the agent to street-level horizontal movement.  Aerial VLN breaks this planar constraint entirely: the UAV navigates in the full $SE(3)$/6-DoF configuration space with three translational axes and three rotational axes, which dramatically increases the dimensionality of planning and control~\cite{AGGARWAL2020270, Mohsan2023UnmannedAV}. The higher state dimensionality of Aerial VLN fundamentally shapes how instructions are expressed and how visual grounding must be performed.

\subsubsection{Viewpoint Dynamics}

In indoor VLN, the camera observes scenes from a roughly constant height and orientation varies only through horizontal rotation. Street-level outdoor VLN similarly maintains a fixed camera height, with viewpoint changes limited to panning. In both settings, the visual appearance of landmarks remains relatively stable across consecutive observations, allowing cross-modal alignment models to rely on consistent visual--semantic associations~\cite{li2025multimodalalignmentfusionsurvey}. Aerial VLN disrupts this stability. As a UAV ascends, descends, banks, and rotates through 3D space, the same physical landmark, for example, a building, or a road intersection, can produce drastically different image inputs. This nonlinear viewpoint variation causes scale distortion, aspect ratio changes, and occlusion patterns that defeat alignment methods designed for stable ground-level perspectives~\cite{qiaoEnhancingVisualAligning2025, wangRealisticUAVVisionlanguage2024}.

\subsubsection{Spatial Scale and Instruction Complexity}

Indoor VLN operates at R2R scale, with typical trajectories spanning 5--15 meters~\cite{gaoAdaptiveZoneawareHierarchical2023, songLonghorizonVisionlanguageNavigation2025a}, and instructions averaging approximately 29 words in R2R~\cite{anderson2018vision}.  Outdoor ground VLN extends to street-block or neighborhood scale, with trajectories of hundreds of meters and instructions of roughly 80 words in benchmarks like Touchdown~\cite{chenTOUCHDOWNNaturalLanguage2019}.  Aerial VLN pushes further to city-level scale. Trajectories in AerialVLN~\cite{liuAerialVLNVisionandlanguageNavigation2023} average over 200 meters with instructions exceeding 80 words, and recent benchmarks like OpenFly~\cite{gaoOpenFlyComprehensivePlatform2025a} feature even longer trajectories across diverse urban, campus, and historical environments. The spatial expansion makes instructions become structurally complex, interleaved spatial-temporal-conditional navigation commands. Decomposing these lengthy instructions into executable sub-goals is itself a significant research challenge~\cite{zhouStructuredInstructionParsing2025, liuAerialVLNVisionandlanguageNavigation2023}. The long-horizon Aerial VLN tasks also mean that the UAV must maintain coherent alignment between its cumulative trajectory and the full instruction over extended periods, placing heavy demands on memory, spatial reasoning, and progress tracking~\cite{zhangDemoAbstractEmbodied2024}.

\subsubsection{Computational Constraints}
UAVs carry limited payload, which restricts onboard compute to lightweight GPUs, and flight time is also battery-limited. Most current Aerial VLN methods sidestep these constraints by offloading computation to a ground station~\cite{zhangCityNavAgentAerialVisionandlanguage2025, sautenkovUAVCodeAgentsScalableUAV2025}, but this reliance on communication links introduces latency and fragility that ultimately must be addressed for real-world deployment~\cite{chenTypeFlyFlyingDrones2024}. We return to this issue in Section~\ref{sec:openproblems}.

\medskip

In summary, Aerial VLN inherits the fundamental cross-modal alignment challenge of indoor and outdoor VLN, i.e. the need to ground language in visual perception to produce actions, but substantially raises the difficulty along some aspects: degrees of freedom, viewpoint stability, spatial scale, instruction complexity, and deployment constraints.  These compounded challenges explain why ground VLN methods can't be straightforwardly transferred to aerial platforms~\cite{wangRealisticUAVVisionlanguage2024, XuEmbodiedNavigation} and motivate the dedicated methodological development surveyed in Section~\ref{sec:methods}.

\begin{figure*}[t]  
    \centering
    \includegraphics[width=0.8\linewidth]{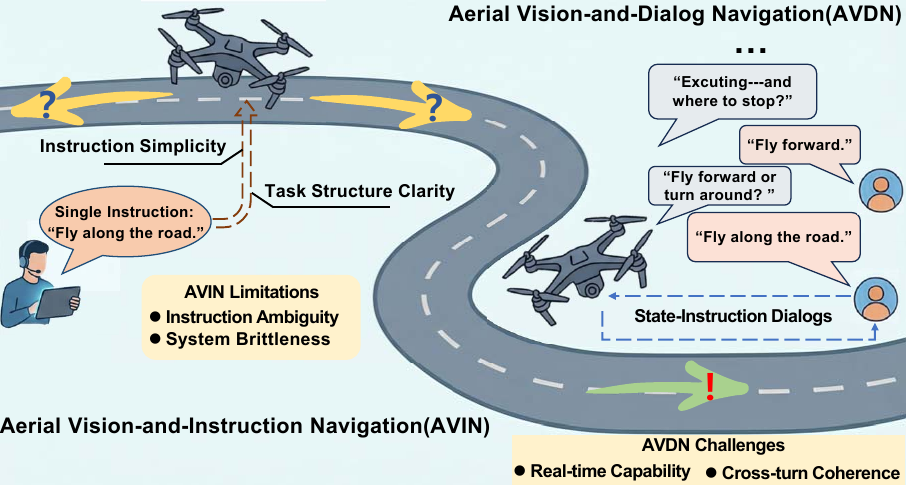}  
    \caption{Interaction paradigms of Aerial VLN: Aerial vision-and-instruction navigation (AVIN) vs. aerial vision-and-dialog navigation (AVDN).}
    \label{fig:Interaction_Paradigms}
\end{figure*}

\subsection{Interaction Paradigms}\label{sec:interaction}
Aerial VLN methods differ in how the operator communicates with the UAV during a navigation episode. Two interaction paradigms have emerged in Aerial VLN methods, distinguished by whether the language input is provided once at the start or evolves through ongoing dialog, as illustrated in Fig. \ref{fig:Interaction_Paradigms}. We define them here as foundational concepts because the choice of paradigm shapes the observation space, the demands on the navigation policy, and the applicable evaluation criteria across all method categories.

\subsubsection{Aerial Vision-and-Instruction Navigation (AVIN)}

In the AVIN paradigm, the operator provides a single, complete natural language instruction $\mathcal{L}$ before the episode begins, and no further linguistic input is available during navigation. The instruction typically describes a route from the starting position to the goal, specifying a sequence of landmarks, turns, and altitude changes that the UAV should follow. Formally, the language component of the observation (Eq.~(\ref{eq:observation})) remains constant throughout the episode: $\mathcal{L}_t = \mathcal{L}$ for all $t$.  The UAV must parse the full instruction, decompose it into an internal plan, and execute that plan using only its visual observations and state estimates as feedback.

AVIN is the dominant paradigm in current Aerial VLN research~\cite{liuAerialVLNVisionandlanguageNavigation2023, leeCityNavLargescaleDataset2025, gaoOpenFlyComprehensivePlatform2025a, zhaoAerialVisionandlanguageNavigation2025, blukisFollowingHighlevelNavigation2018}. Its appeal lies in its simplicity and clear task structure: success or failure is determined entirely by whether the UAV can ground and execute a fixed instruction.  However, this simplicity comes at a cost.  Because the instruction is issued once and cannot be revised, any ambiguity, error, or mismatch between the instruction and the UAV's perception must be resolved by the agent alone.  If the UAV misinterprets a landmark reference or loses track of its progress along the instruction, there is no mechanism for correction. This brittleness becomes particularly acute in Aerial VLN, where lengthy, structurally complex instructions and viewpoint variation can cause UAV navigation to fail~\cite{liuAerialVLNVisionandlanguageNavigation2023, zhouStructuredInstructionParsing2025}.

\subsubsection{Aerial Vision-and-Dialog Navigation (AVDN)}

In the AVDN paradigm, navigation is guided by a multi-turn dialog between the UAV and the operator. Rather than receiving a single monolithic instruction, the UAV obtains an initial directive and can subsequently receive clarifications, corrections, or new sub-instructions based on the evolving navigation context. Formally, the language input at time $t$ is augmented by a dialog history $\mathcal{D}_t = (d_1, d_2, \ldots, d_t)$, where each dialog turn $d_i$ contains an operator utterance:
\begin{equation}
    o_t = (\mathcal{V}_t, \hat{\mathbf{p}}_t, \mathcal{L}, \mathcal{D}_t)
    \label{eq:observation_avdn}
\end{equation}

The AVDN paradigm was introduced to Aerial VLN by Fan et al.~\cite{fanAerialVisionanddialogNavigation2023a} with AVDN dataset. The key advantage of AVDN over AVIN is its capacity for progressive ambiguity resolution: this dialog-based closed-loop mechanism enables operator guidance during unexpected scenes or unclear instructions, more faithfully mirroring real-world HRI in Aerial VLN.

However, AVDN introduces its own challenges. The navigation policy must process the growing dialog history in addition to visual observations, requiring the model to maintain cross-turn coherence and extract the operator's evolving intent from potentially redundant or contradictory exchanges.  Current AVDN methods~\cite{fanAerialVisionanddialogNavigation2023a, suTargetgroundedGraphawareTransformer2023, suLearningFinegrainedAlignment, qiaoEnhancingVisualAligning2025} address this by encoding the dialog history alongside visual and state features, but they operate on pre-collected dialog transcripts rather than generating dialog in real time. The gap to conduct live interactive dialog and integrate the response into ongoing navigation remain largely open.

\subsubsection{Relationship Between Paradigms and Methods}

It is important to note that AVIN and AVDN define the interaction interface, not the method architectures in Section~\ref{sec:methods}. In practice, the vast majority of current methods are evaluated under the AVIN paradigm, because AVIN benchmarks are more numerous, better standardized, and easier to evaluate. AVDN methods remain a small but growing subset of Aerial VLN, and their evaluation requires additional metrics, such as dialog efficiency and instruction completion rate across turns that are not yet standardized.

Fig.~\ref{fig:Interaction_Paradigms} illustrates the information flow under both paradigms. Throughout the method review in Section~\ref{sec:methods}, we indicate which interaction paradigm each method targets, treating AVIN and AVDN as a cross-cutting label rather than a structural division.

\section{Methods for Aerial VLN}\label{sec:methods}

\begin{figure*}[t]  
    \centering
    \includegraphics[width=1\linewidth]{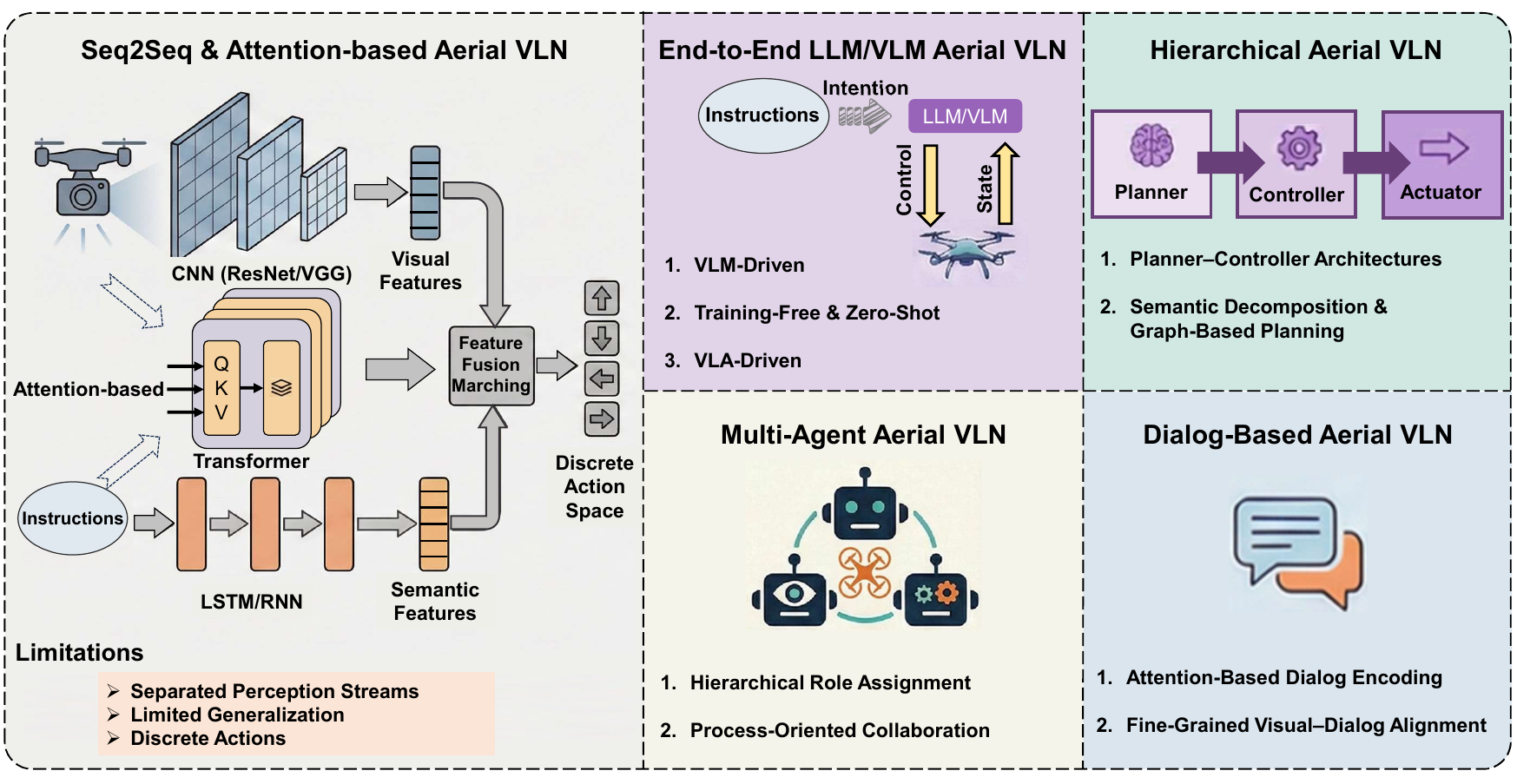}  
    \caption{The taxonomy of Aerial VLN methods.}
    \label{fig:Methods_Taxonomy}
\end{figure*}

This section reviews Aerial VLN methods under a taxonomy organized by the architecture of the navigation policy $\pi$ (Eq.~(\ref{eq:policy})).  We identify five categories.  \textbf{Sequence-to-sequence and attention-based methods} (Section~\ref{sec:seq2seq}) encode instructions and visual observations via task-specific encoders and decode discrete action sequences through cross-modal feature matching.  \textbf{End-to-end LLM/VLM methods} (Section~\ref{sec:e2e}) replace task-specific encoders with large pre-trained models that directly map multimodal inputs to navigation actions.  \textbf{Hierarchical methods} (Section~\ref{sec:hierarchical}) decouple high-level semantic planning that is performed by LLMs from low-level flight control executed by traditional controllers.  \textbf{Multi-agent methods} (Section~\ref{sec:multiagent}) distribute the navigation task across multiple collaborating LLM-based agents, each responsible for a distinct functional role.  \textbf{Dialog-based navigation methods} (Section~\ref{sec:dialog}) process multi-turn dialog histories to extract evolving navigation intent under the AVDN paradigm (Section~\ref{sec:interaction}).  Fig.~\ref{fig:Methods_Taxonomy} illustrates this taxonomy.

\subsection{Sequence-to-Sequence and Attention-Based Methods}
\label{sec:seq2seq}
The earliest Aerial VLN methods adopted the sequence-to-sequence (Seq2Seq) paradigm inherited from indoor VLN~\cite{anderson2018vision, friedSpeakerfollowerModelsVisionandlanguage2018}: natural language instructions and visual observations are encoded into fixed-dimensional feature vectors by separate encoders, fused through a matching or attention mechanism, and decoded into a sequence of navigation actions.  These methods established the foundational formulation of Aerial VLN as a tractable sequential decision-making problem. But because methods in this category predate the integration of LLM/VLM into Aerial VLN, it comes at the cost of limited model generalization and methods are confined to task-specific learning.

\subsubsection{Instruction Parsing and Semantic Mapping}

The earliest work on language-guided UAV navigation~\cite{huangNaturalLanguageCommand2010} relied on hand-crafted natural language processing (NLP) modules to parse instructions into discrete semantic units: actions, landmarks, and spatial relations, which were then grounded on a map through probabilistic models. While pioneering, this approach required manual annotation and rule engineering, limiting its scalability and generalization.

Subsequent work shifted toward learned representations. Misra et al.~\cite{misraMappingInstructionsActions2018} proposed decomposing Aerial VLN into two stages: predicting visual goals location from the instruction, and generating actions to reach goals. LANI dataset, one of the first outdoor Aerial VLN datasets, was constructed though limited to simplified 2D environments. Blukis et al.~\cite{blukisMappingNavigationInstructions, blukisLearningMapNatural} advanced this line by projecting image features and instruction features onto a semantic map from the UAV's perspective. A convolutional policy network then consumed this map to predict position-visitation distributions for visit or stop of the UAV. The navigation actions were generated through IL~\cite{rossReductionImitationLearning2011} and RL~\cite{blukisLearningMapNatural}, producing continuous low-level velocity commands for real-time control. This was among the first Aerial VLN methods to output continuous actions rather than discrete primitives, though the approach was evaluated only in simplified simulation environments with geometric structures and limited visual complexity.

\subsubsection{Cross-Modal Attention and Dynamic Perception}

As Aerial VLN tasks grew in visual and instructional complexity, attention mechanisms became central to aligning language and vision features. Transformer architectures are integrated to enhance the perceptual capacity and contextual awareness of Aerial VLN\cite{dingHistoryenhancedTwostageTransformer2025, XuMultimodalLearning}. AerialVLN/AerialVLN-S benchmark\cite{liuAerialVLNVisionandlanguageNavigation2023} established the first standardized baselines for aerial instruction-following and introduced a cross-modal attention mechanism specifically designed for lengthy aerial instructions. This approach segments a long instruction into multiple sub-instructions and aligns each sub-instruction with the corresponding path segment through temporal attention. The visual encoder uses a ResNet~\cite{heDeepResidualLearning2016a} backbone to extract image features, and the language encoder uses a LSTM~\cite{6795963} to produce contextualized word embeddings. Actions are selected from a discrete set with 4-DoF control (forward/backward, left/right, up/down, and rotation).

To model dynamically evolving scene information, the dual-branch dynamic perception and interaction framework (DBDP) for Aerial VLN \cite{wangDualbranchDynamicPerception2025} adopted a functionally decoupled design to capture dynamic features. A CNN-LSTM branch captures temporal visual dynamics while an attention-based branch extracts directional spatial cues. This dual-branch architecture enhanced dynamic understanding for continuous UAV navigation decisions.
The bird’s eye view (BEV) grid map from multi-view observation skybox \cite{zhaoAerialVisionandlanguageNavigation2025} was designed for Aerial VLN. The correlative model selected viewpoints compliant with the instruction processed by pre-trained BERT~\cite{devlinBERTPretrainingDeep2019}, and aligned instructions with the BEV grid map through cross-modal attention to execute corresponding navigation movements.

\subsubsection{Safety-Aware and Control-Integrated Methods}

A distinct thread within this category addresses the gap between high-level action prediction and safe, physically realizable flight. The adaptive safety margin algorithm (ASMA)\cite{sanyalASMAUnderlinetextAdaptiveUnderlinetextSafety2025} integrates control barrier functions (CBFs) with model predictive control (MPC) to dynamically adjust flight trajectories for collision avoidance during VLN execution. This represents one of the few Aerial VLN works that explicitly addresses flight safety within the navigation loop. SINGER~\cite{adangSINGEROnboardGeneralist2025} employed IL to acquire a flight policy, trained on expert trajectories generated by an RRT* planner and an MPC controller. This control-integrated Aerial VLN policy is designed for real-time execution on aerial platforms.


Table~\ref{tab:seq2seq_methods} summarizes the methods in this category along the four comparative dimensions. Several shared limitations are evident. First, the discrete action space adopted by most methods in this category (with the exception of Blukis et al.~\cite{blukisMappingNavigationInstructions, blukisLearningMapNatural}) is a significant simplification that constrains the UAV to coarse and grid-like movements. This discretization was inherited from indoor VLN and is poorly suited to the continuous 6-DoF dynamics of real UAV flight.
Second, these methods exhibit limited generalization across environments, because the visual and language encoders are trained from task-specific datasets. The absence of large-scale pre-training, which would later be addressed by LLM-centric methods, is the fundamental bottleneck.
Third, cross-modal alignment is brittle under the viewpoint variation characteristic of aerial navigation~\cite{mishra2025aermani}. Aerial viewpoint changes cause the same landmark to produce vastly different feature representations. Attention mechanisms mitigate this to some extent, but the core challenge of visual features inconsistency still persists. These limitations motivated the transition toward methods that leverage the semantic reasoning, world knowledge, and zero-shot generalization capabilities of pre-trained LLMs, as discussed in the following subsections.

\begin{table*}[t]
\centering
\caption{Summary of sequence-to-sequence and attention-based Aerial VLN methods.}
\label{tab:seq2seq_methods}
\setlength{\tabcolsep}{2pt}
\small
\begin{tabular}{l c c c c}
\toprule
\textbf{Method} & \textbf{Input Repr.} & \textbf{Model Role} & \textbf{Action Format} & \textbf{Benchmarks} \\
\midrule
Huang et al.~\cite{huangNaturalLanguageCommand2010} & RGB + parsed NL & Rule-based parser & Discrete & Custom \\
Misra et al.~\cite{misraMappingInstructionsActions2018} & RGB + NL & Goal predictor + actor & Discrete & LANI \\
Blukis et al.~\cite{blukisMappingNavigationInstructions,blukisLearningMapNatural} & Semantic map & Visitation predictor & Continuous ($v$, $\omega$) & Custom (simplified) \\
AerialVLN~\cite{liuAerialVLNVisionandlanguageNavigation2023} & RGB-D + NL (sub-instr.) & Cross-modal decoder & Discrete (4-DoF) & AerialVLN, AerialVLN-S \\
DBDP~\cite{wangDualbranchDynamicPerception2025} & RGB seq. + NL & Dual-branch encoder--decoder & Discrete & AerialVLN \\
Zhao et al.~\cite{zhaoAerialVisionandlanguageNavigation2025} & Multi-view skybox + NL & BEV selector + BERT & Discrete & AerialVLN \\
ASMA~\cite{sanyalASMAUnderlinetextAdaptiveUnderlinetextSafety2025} & RGB + NL + UAV state & Waypoint predictor + CBF-MPC & Continuous (waypoints) & Custom \\
SINGER~\cite{adangSINGEROnboardGeneralist2025} & RGB + NL + UAV state & IL policy (RRT*+MPC expert) & Continuous & Custom \\
\bottomrule
\end{tabular}
\end{table*}

\subsection{End-to-End LLM/VLM Methods}\label{sec:e2e}
End-to-end methods replace the task-specific encoders and decoders of Section~\ref{sec:seq2seq} with pre-trained LLMs or VLMs~\cite{liuVisualInstructionTuning, chuVisionLLaMAUnifiedLLaMA, wangInternVL35AdvancingOpensource2025} that directly map instructions, UAV state, and environmental perception into navigation actions. The central promise of this category is that the semantic reasoning, world knowledge, and generalization capabilities acquired during large-scale pre-training can compensate for the limited scale and diversity of Aerial VLN datasets~\cite{zhaoSurveyLargeLanguage2025, naveedComprehensiveOverviewLarge2024, minaeeLargeLanguageModels2025}. However, a persistent tension runs through these methods: most still output discrete actions despite operating with LLMs capable of far richer representations, because continuous UAV control remains difficult to learn end-to-end from language supervision alone.

\subsubsection{VLM-Driven Discrete Action Prediction}

The dominant paradigm within this category uses VLMs to reason over multimodal inputs and output discrete actions.  NavAgent~\cite{liuNavAgentMultiscaleUrban2024} is the first urban UAV embodied navigation model driven by a VLM. It constructs a multi-scale environmental representation---topological maps at the global scale, panoramic images at the medium scale, and fine-grained landmark descriptions at the local scale---and feeds these alongside navigation instructions into the VLM to reason and generate discrete actions.
FlightGPT~\cite{caiFlightGPTGeneralizableInterpretable2025} introduces a chain-of-thought (CoT) reasoning mechanism and supervised fine-tuning (SFT) for action prediction. Then combined with RL, it maps the reasoning process to discrete action sequences.
The LLM-centric Aerial VLN framework based on semantic-topo-metric representation (STMR) \cite{gaoAerialVisionandlanguageNavigation2024, gaoExploringSpatialRepresentation2025} is designed to extract and project instruction-relevant semantic masks onto a top-down map. The spatial information including topology, semantics, and metrics is then transformed into a structured matrix for input of the LLM. This matrix as the explicit spatial representation mitigates the limited native ability of LLMs to reason about precise metric and topological relationships from raw images~\cite{gaoExploringSpatialRepresentation2025}.
GeoNav~\cite{xuGeoNavEmpoweringMLLMs2025} extends the spatial reasoning approach by constructing a dual-scale spatiotemporal perception memory: a cognitive map for global path planning and a hierarchical scene graph for local target localization. Both representations are integrated into a multimodal LLM (MLLM) via a multimodal CoT prompting mechanism. The MLLM outputs discrete action sequences at different granularities, enabling progressive navigation from macro-level route selection to local goal approach. This dual-scale architecture offers a solution to the long-range dependency problem.

Several additional methods operate within this discrete-action VLM paradigm.  LogisticsVLN~\cite{zhangLogisticsVLNVisionlanguageNavigation2025}, targeting low-altitude terminal delivery, uses multiple lightweight VLMs to address direction selection, target detection, and floor estimation in separate modules, ultimately producing discrete action sequences. UAV-ON~\cite{xiaoUAVONBenchmarkOpenworld2025} replaces path-description-guided navigation with object-description-guided navigation, using an Aerial ObjectNav Agent (AOA) module containing an LLM to generate actions based on UAV state and four-directional environmental perception. SA-GCS \cite{caiSAGCSSemanticawareGaussian2025} employs semantic-aware Gaussian curriculum scheduling, a strategy optimized via RL, to enhance the decision-making generalization of VLMs in Aerial VLN tasks.  OpenVLN~\cite{linOpenVLNOpenworldAerial2025} optimizes VLM updates via a novel reward function using rule-based RL, enabling effective fine-tuning with minimal data and significantly enhancing long-horizon VLN capability in complex aerial environments.

\subsubsection{Training-Free and Zero-Shot Approaches}

A distinct thread within end-to-end methods explores whether pre-trained VLMs can navigate without any task-specific training, relying solely on zero-shot prompting. SPF \cite{huSeePointFly2025} improves the low precision of zero-shot action prediction by decomposing navigation problem. the VLM is used to annotate 2D waypoints on the input image based on the instruction. These waypoints are then converted into 3D displacement vectors through geometric projection and subsequently decomposed into angle and throttle commands.
CoDrone~\cite{chenCoDroneAutonomousDrone2025} employs a split architecture with edge and cloud foundation models for end-to-end navigation. This design reduces onboard computational overhead while maintaining navigation stability, making it one of the few methods that explicitly addresses the deployment constraints identified in Section~\ref{sec:context}.

\subsubsection{Vision-Language-Action (VLA) Methods}

A growing subset of end-to-end methods adopts the vision-language-action (VLA) framework~\cite{Li2025IBAMG, zitkovichRT2VisionlanguageactionModels2023}, which unifies visual perception, semantics reasoning, and action generation. VLA methods are distinguished from the discrete-action VLM approaches above by their ambition to produce continuous or spatially grounded action outputs rather than selecting from a finite action set.

UAV-VLA~\cite{sautenkovUAVVLAVisionlanguageactionSystem2025} and its extension UAV-VLPA*~\cite{sautenkovUAVVLPAVisionlanguagepathactionSystem2025} leverage satellite imagery to generate path-action sets for UAVs through a combination of VLM visual analysis and GPT-based instruction processing. With reference paths generated by VLMs, LLMs perform action reasoning to achieve spatially grounded action generation.
GRaD-Nav++~\cite{chenGRADNAVVisionlanguageModel2026} employs a Mixture-of-Experts (MoE) strategy~\cite{huang2025mentor} to train VLA models via RL in a 3D Gaussian Splatting (3DGS) rendered environment~\cite{Kerbl3DGaussianSplatting}. The entire pipeline is formulated as a POMDP (Section~\ref{sec:formulation}), with the VLA model serving as the policy. And the use of 3DGS for rendering narrows the sim-to-real gap.
UAV-Flow Colosseo~\cite{wangUAVflowColosseoRealworld2025} establishes a real-world benchmark for evaluating VLA models in Aerial VLN, and introduces UAV IL based on VLA for short-range, reactive flight behaviors.


Table~\ref{tab:e2e_methods} summarizes end-to-end methods along the five comparative dimensions. And some characteristics are as followed:
First is the persistent dominance of discrete actions. Despite the reasoning capacity of LLMs/VLMs, the majority of end-to-end methods still output actions from a set of directional primitives. This is partly a practical choice: discrete action prediction aligns naturally with the token-generation paradigm of LLMs/VLMs, and also reflects the difficulty of mapping continuous control. VLA methods represent the most direct attempt to bridge this gap, but still remain in early stages for Aerial VLN.
The second is the growing role of structured spatial representations as LLM inputs. Methods like STMR~\cite{gaoAerialVisionandlanguageNavigation2024}, GeoNav~\cite{xuGeoNavEmpoweringMLLMs2025}, and NavAgent~\cite{liuNavAgentMultiscaleUrban2024} construct structured spatial representations (semantic maps, schematic cognitive map, scene topology map) that convert visual information into a format the LLM can reason about more effectively.
The third is the diversity of training strategies. Methods in this category span the spectrum of zero-shot prompting~\cite{huSeePointFly2025}, SFT~\cite{liuNavAgentMultiscaleUrban2024}, multi-stage SFT+RL pipelines~\cite{caiFlightGPTGeneralizableInterpretable2025, linOpenVLNOpenworldAerial2025} and VLA training~\cite{chenGRADNAVVisionlanguageModel2026}. No single training paradigm has emerged as clearly dominant, and the optimal strategy likely depends on the expert demonstrations, action space, data structure and so on.

\begin{table*}[t]
\centering
\caption{Summary of end-to-end LLM/VLM Aerial VLN methods.}
\label{tab:e2e_methods}
\setlength{\tabcolsep}{4pt}
\small
\begin{tabular}{l c c c c c}
\toprule
\textbf{Method} & \textbf{Input Repr.} & \textbf{Model Role} & \textbf{Action Format} & \textbf{Training} & \textbf{Benchmarks} \\
\midrule
NavAgent~\cite{liuNavAgentMultiscaleUrban2024} & Multi-scale RGB + NL & VLM action predictor & Discrete & SFT & AerialVLN \\
FlightGPT~\cite{caiFlightGPTGeneralizableInterpretable2025} & RGB + NL + CoT prompt & VLM with CoT & Discrete & SFT + RL & AerialVLN \\
Gao et al.~\cite{gaoAerialVisionandlanguageNavigation2024} & STMR matrix + NL & LLM with CoT & Discrete & Zero-shot & AerialVLN-S, OpenFly \\
GeoNav~\cite{xuGeoNavEmpoweringMLLMs2025} & Cognitive map + RGB + NL & MLLM with CoT & Discrete & SFT & CityNav \\
LogisticsVLN~\cite{zhangLogisticsVLNVisionlanguageNavigation2025} & RGB + NL & Multiple VLMs & Discrete & SFT & Custom (delivery) \\
UAV-ON~\cite{xiaoUAVONBenchmarkOpenworld2025} & 4-dir RGB + NL + state & LLM action predictor & Discrete & SFT & UAV-ON \\
Cai et al.~\cite{caiSAGCSSemanticawareGaussian2025} & RGB + NL & VLM with GCS & Discrete & RL & AerialVLN \\
OpenVLN~\cite{linOpenVLNOpenworldAerial2025} & RGB + NL & VLM with rule-based RL & Discrete & RL & AerialVLN, OpenFly \\
Hu et al.~\cite{huSeePointFly2025} & RGB + NL & VLM waypoint annotator & Continuous & Zero-shot & Custom \\
CoDrone~\cite{chenCoDroneAutonomousDrone2025} & RGB + NL & Edge+cloud VLM & Discrete & SFT & Custom \\
UAV-VLA~\cite{sautenkovUAVVLAVisionlanguageactionSystem2025} & Satellite RGB + NL & VLM+GPT path planner & Waypoints & SFT & UAV-VLA \\
GRaD-Nav++~\cite{chenGRADNAVVisionlanguageModel2026} & 3DGS-rendered RGB + NL & VLA (MoE) & Continuous & RL & Custom (3DGS) \\
UAV-Flow~\cite{wangUAVflowColosseoRealworld2025} & RGB + NL & VLA (IL) & Continuous & IL & UAV-Flow (real-world) \\
\bottomrule
\end{tabular}
\end{table*}

\subsection{Hierarchical Methods}\label{sec:hierarchical}
Hierarchical methods explicitly decouple the Aerial VLN problem into two layers: a high-level planner, typically powered by LLMs/VLMs, that performs task decomposition and semantic reasoning~\cite{Waga2025ASO}; and a low-level executor that translates the planner's output into physically realizable flight commands using established UAV control controllers. This separation offers a key practical advantage: the high-level planner can leverage the full reasoning power of LLMs/VLMs without being constrained by real-time control loop requirements, while the low-level executor inherits the robustness of mature flight control methods. Among all Aerial VLN method categories, hierarchical methods are the most directly compatible with existing UAV autonomy stacks, making a promising pathway toward real-world deployment.

\subsubsection{Planner--Controller Architectures}

SkyVLN~\cite{liSkyVLNVisionandlanguageNavigation2025} exemplifies the hierarchical paradigm.  Its high-level planner is an LLM-based motion command generator that takes navigation prompts as inputs, and outputs structured thoughts and actions in natural language.  The low-level executor uses nonlinear model predictive control (NMPC) to track the planned trajectory, providing dynamic obstacle avoidance and precise trajectory following.  The planner-controller interface takes the form of motion commands that the NMPC module converts into continuous velocity and attitude references.
VLFly~\cite{zhangGroundedVisionlanguageNavigation2025} adopts a different interface design. Its high-level layer uses an LLM-based instruction encoder to reformulate the raw instruction into structured prompts, and a VLM-powered goal retriever for zero-shot navigation target detection.  The low-level layer employs a waypoint planner based on ViNT~\cite{shahViNTFoundationModel2023} to generate continuous velocity commands, executing monocular visual navigation.  The planner-controller interface is the detected visual target, which preserves metric information that would be lost in instruction communication.
AirStar~\cite{wangHiAirStarGuide2025} introduces a library-based interaction design.  Its LLM task planner parses the instruction and selects appropriate navigation strategies from a predefined navigation library and UAV behaviors from a skill library.  The execution layer combines the A* algorithm for global path planning with the Ego-Planner~\cite{zhouEGOPlannerESDFFreeGradientBased2021} for local trajectory optimization. The planner-controller interface in AirStar is a structured task specification, and this library-based approach offers high modularity, but it constrains the system to the predefined set of behaviors in the libraries.

\subsubsection{Semantic Decomposition and Graph-Based Planning}

CityNavAgent~\cite{zhangCityNavAgentAerialVisionandlanguage2025} combines LLM-based instruction decomposition with graph-based spatial planning. The LLM decomposes the complex natural language instruction into subgoals at different semantic levels, and the UAV then navigates between subgoal nodes based on the topological map and graph search algorithm. The planner-controller interface is a sequence of topological graph nodes. CityNavAgent was evaluated on the CityNav benchmark \cite{leeCityNavLargescaleDataset2025}, and its hierarchical semantic planning demonstrated particular effectiveness on long instructions.
OpenUAV~\cite{wangRealisticUAVVisionlanguage2024} presents a complete hierarchical architecture for complex and long-range Aerial VLN. Its high-level layer employs a MLLM to integrate multi-view imagery and natural language instructions, producing a coarse global plan in the form of a distant target pose.  The low-level layer then performs local path. OpenUAV supports continuous 6-DoF flight control, which is achieved through the combination of MLLM-based global reasoning and vision-based local planning. The planner-controller interface is the target pose $(\mathbf{x}^*, \mathbf{q}^*) \in SE(3)$. OpenUAV was evaluated on 22 scenes with 12,000 6-DoF trajectories and demonstrated strong performance on long-range tasks.

\begin{table*}[t]
\centering
\caption{Summary of hierarchical Aerial VLN methods.  The interface column describes the intermediate representation between the high-level planner and the low-level controller.}
\label{tab:hierarchical_methods}
\setlength{\tabcolsep}{4pt}
\small
\begin{tabular}{l c c c c}
\toprule
\textbf{Method} & \textbf{High-Level Planner} & \textbf{Low-Level Controller} & \textbf{Interface} & \textbf{Action Format} \\
\midrule
SkyVLN~\cite{liSkyVLNVisionandlanguageNavigation2025} & LLM + WPO & NMPC & Motion commands (text) & Continuous (via NMPC) \\
VLFly~\cite{zhangGroundedVisionlanguageNavigation2025} & LLM encoder + VLM & ViNT waypoint planner & Visual target (image region) & Continuous ($(v, w)$) \\
AirStar~\cite{wangHiAirStarGuide2025} & LLM task planner & A* + Ego-Planner & Task spec (mode + skill) & Continuous (trajectory) \\
CityNavAgent~\cite{zhangCityNavAgentAerialVisionandlanguage2025} & LLM + global memory & Graph search + local controller & Topological graph nodes & Discrete (graph hops) \\
OpenUAV~\cite{wangRealisticUAVVisionlanguage2024} & MLLM & Vision-based local planner & Target pose $\in SE(3)$ & Continuous (6-DoF) \\
\bottomrule
\end{tabular}
\end{table*}


Table~\ref{tab:hierarchical_methods} summarizes hierarchical methods along the four reporting dimensions. 
First, the planner-controller interface is the core design choice, and its form has significant implications.  Text-based interfaces (SkyVLN~\cite{liSkyVLNVisionandlanguageNavigation2025}) are the most natural for LLM planners but sacrifice spatial precision.  Visual grounding interfaces (VLFly~\cite{zhangGroundedVisionlanguageNavigation2025}) preserve metric information but require a robust target detection module.  Topological graph interfaces (CityNavAgent~\cite{zhangCityNavAgentAerialVisionandlanguage2025}) provide structured spatial reasoning but depend on the availability and quality of the graph representation.  Target pose interfaces (OpenUAV~\cite{wangRealisticUAVVisionlanguage2024}) offer the most direct coupling to 6-DoF flight control but demand accurate global reasoning from the high-level planner. No single interface design dominates, the optimal choice depends on the compatibility,  navigation scale, instruction complexity and available environmental representations.
Second, hierarchical methods achieve the natural integration with existing UAV autonomy stacks. This contrasts with end-to-end methods without the safety guarantees that come from modular, well-tested control layers.  For deployment-oriented research, this compatibility is also a substantial advantage.
Additionally hierarchical methods exhibit feasible in matching planning and control different latency. Different interfaces are designed to bridge the gap between the low-frequency operation of high-level LLM planners and the high-frequency demands of low-level controllers. but the fundamental tension between planning frequency and action responsiveness remains an open design challenge.

\subsection{Multi-Agent Methods}\label{sec:multiagent}
Multi-agent methods distribute the Aerial VLN task across multiple collaborating LLM-based agents, each assigned a distinct functional role. Research on Multi-agent methods is still in nascent stage, with only a handful of studies published to date, but the architectural ideas are instructive and point toward a potentially productive research direction.

The hierarchical role assignment method UAV-CodeAgents~\cite{sautenkovUAVCodeAgentsScalableUAV2025} presents a scalable Aerial VLN framework that integrates a multi-agent system with the reactive thinking loop (ReAct) paradigm. This framework combines LLMs and VLMs to generate UAV flight trajectories from inputs of satellite imagery and natural language instructions. UAVCodeAgents comprises two specialized agents, an airspace manager agent and an UAV agent. The airspace manager agent is responsible for high-level semantic perception and adaptive planning, while the UAV agent handles low-level execution and real-time feedback. Alternatively, Inspired by multi-agent methods in ground VLN, multi-agent system can be structured around the core procedural nodes of Aerial VLN. The process-oriented collaboration method MMCNav~\cite{zhangMMCNavMLLMempoweredMultiagent2025} conducts VLN tasks by constructing specialized agents of observation, planning, execution, and feedback. The interaction and collaborative decision-making among these agents simulates the navigation process, leading to systematic improvements in multi-modal perception and the reliability of planning decisions. Multi-agent Aerial VLN methods significantly enhances the UAV’s capacity for deep scene understanding and complex task decomposition. Furthermore, the structured interactions among agents foster greater overall system intelligence and operational reliability.

Table~\ref{tab:multiagent_methods} summarizes and compares the two multi-agent methods.
The two methods represent complementary organizational strategies, i.e., vertical manager-executor pattern and horizontal process pipeline pattern. The vertical design provides clear authority and rapid strategic adaptation, while the horizontal design enables rational execution logic and error correction mechanism. But the empirical case for multi-agent over single-agent remains thin.  It is difficult to determine whether the observed benefits stem from the multi-agent architecture itself or simply from the additional model capacity and structured prompting that multi-agent designs introduce. Additionally, multi-agent methods face a scalability question that is particularly relevant for aerial platforms.  Each additional agent introduces inference overhead, inter-agent communication latency, and coordination complexity.  On a resource-constrained UAV platform (Section~\ref{sec:context}), these costs are non-trivial.  Future multi-agent Aerial VLN research need to demonstrate that collaboration improves navigation quality under additional computational and latency costs.

\begin{table*}[t]
\centering
\caption{Summary of multi-agent Aerial VLN methods.}
\label{tab:multiagent_methods}
\small
\begin{tabular}{l c c}
\toprule
& \textbf{UAV-CodeAgents}~\cite{sautenkovUAVCodeAgentsScalableUAV2025} & \textbf{MMCNav}~\cite{zhangMMCNavMLLMempoweredMultiagent2025} \\
\midrule
\textit{Agent roles} & Airspace manager + UAV agent & Observation + Planning + Execution + Feedback \\
\textit{Organization} & Vertical (manager--executor) & Horizontal (process pipeline) \\
\textit{Communication} & ReAct loop (text-based) & Sequential pipeline + feedback loop \\
\textit{Model backbone} & LLM + VLM (both agents) & MLLM (all agents) \\
\textit{Action format} & Trajectory waypoints & Discrete actions \\
\textit{Training} & Zero-shot (ReAct prompting) & SFT \\
\textit{Benchmarks} & Custom (satellite + NL) & VLN benchmarks \\
\textit{Self-correction} & Via manager re-planning & Via feedback agent \\
\bottomrule
\end{tabular}
\end{table*}

\begin{table*}[t]
\centering
\caption{Summary of dialog-based Aerial VLN methods.  All methods operate under the AVDN paradigm on the dataset of Fan et al.~\cite{fanAerialVisionanddialogNavigation2023a}.}
\label{tab:dialog_methods}
\setlength{\tabcolsep}{4pt}
\small
\begin{tabular}{l c c c c}
\toprule
\textbf{Method} & \textbf{Input Repr.} & \textbf{Model Role} & \textbf{Action Format} & \textbf{Training} \\
\midrule
HAA-Trans.~\cite{fanAerialVisionanddialogNavigation2023a} & Satellite RGB + dialog hist. & Cross-attn.\ encoder--decoder & Waypoints & Supervised \\
Su et al.~\cite{suTargetgroundedGraphawareTransformer2023} & Satellite RGB + dialog hist.\ & Graph-aware grounding + decoder & Waypoints & Supervised \\
Su et al.~\cite{suLearningFinegrainedAlignment} & Satellite RGB + dialog segments & Fine-grained aligner + decoder & Waypoints & Supervised \\
Qiao et al.~\cite{qiaoEnhancingVisualAligning2025} & Satellite RGB + dialog hist.\  & Rotated detector + multi-stage pre-train & Waypoints & Multi-stage pre-train \\
\bottomrule
\end{tabular}
\end{table*}

\subsection{Dialog-Based Navigation Methods}\label{sec:dialog}
The methods in this subsection are designed specifically for the AVDN interaction paradigm (Section~\ref{sec:interaction}). The core technical challenge is to extract coherent navigation intent from a sequence of dialogs. All current AVDN methods for Aerial VLN operate on the dataset introduced by Fan et al.~\cite{fanAerialVisionanddialogNavigation2023a}, which collects asynchronous human-human dialogs from a satellite-view navigation scenario to simulate a commander guides a follower UAV.

\subsubsection{Attention-Based Dialog Encoding}
Fan et al.~\cite{fanAerialVisionanddialogNavigation2023a} established the AVDN task and proposed the human attention aided transformer (HAA-Transformer) as the baseline method. The HAA-Transformer synchronously processes the full dialog history, satellite-view imagery and the UAV's state information. Then decoder predicts specific waypoints on the satellite image.

\subsubsection{Fine-Grained Visual--Dialog Alignment}
Subsequent works by \cite{suTargetgroundedGraphawareTransformer2023, suLearningFinegrainedAlignment} enhanced AVDN navigation performance by strengthening scene understanding and achieving more precise visual-dialog alignment. The target-grounded graph-aware transformer~\cite{suTargetgroundedGraphawareTransformer2023} introduces a structured graph representation to reflect landmarks and spatial relations. Dialogs are then grounded onto this graph through a graph-aware attention mechanism. This structured explicitly encoded spatial relations in the graph topology rather than making it difficult for the attention mechanism to discover implicitly.
In subsequent work, \cite{suLearningFinegrainedAlignment} further advanced fine-grained alignment by learning to associate specific dialog segments with corresponding visual regions at a sub-utterance level. It decomposes dialog turns into constituent referring expressions and aligns each expression with a localized visual region. This decomposition is critical for AVDN because dialog turns in real-world navigation often contain multiple spatial references, and coarse alignment makes navigation unreliable and ineffective.
\cite{qiaoEnhancingVisualAligning2025} improved visual-language alignment and grounding in AVDN through a combination of rotated object detection, multi-stage pre-training and data augmentation. Rotated object detection is particularly relevant for the satellite-view setting. The multi-stage pre-training transforms general visual understanding to task-specific dialog grounding. Geometric data augmentation of the satellite imagery further increases the diversity of training examples, mitigating overfitting to the limited scale of the AVDN dataset.

Table~\ref{tab:dialog_methods} summarizes dialog-based Aerial VLN methods. But there are some limitations in AVDN required further investigation.
One is all existing AVDN methods process a fixed sequence of historical dialog turns to predict waypoints rather than generate dialog in real time. This is a fundamental gap between the current technical reality and the aspiration articulated in the AVDN paradigm definition (Section~\ref{sec:interaction}). Related works in indoor dialog-based navigation~\cite{shridharALFREDBenchmarkInterpreting2020, padmakumarTEAChTaskdrivenEmbodied2022} also offer method references.
The other is all current AVDN methods operate exclusively in the satellite-view setting. While this setting simplifies visual grounding, it is not representative of the UAV-view navigation that most other Aerial VLN methods address.  Extending AVDN to aerial perspectives would significantly increase the difficulty of visual-dialog alignment. 
Addressing these limitations is the way to bring AVDN closer to its original vision of interactive, corrigible aerial navigation guided by natural human-UAV conversation.

\section{Evaluation Infrastructure}\label{sec:infrastructure}
The methods reviewed in Section~\ref{sec:methods} are only as trustworthy as the infrastructure on which they are developed and evaluated.  This section provides a critical assessment of the three pillars of Aerial VLN evaluation infrastructure: datasets (Section~\ref{sec:datasets}), simulation platforms (Section~\ref{sec:simulators}) and evaluation metrics (Section~\ref{sec:metrics}).  For each, we catalog what currently exists and identify where the infrastructure falls short.

\subsection{Datasets}\label{sec:datasets}
Aerial VLN datasets must integrate real-world or photorealistic UAV flight data with natural language instructions, providing the essential training and evaluation foundation for the methods reviewed in Section~\ref{sec:methods}.  We divide existing datasets into two categories: dedicated Aerial VLN datasets and underexploited datasets created for object detection or semantic segmentation tasks with potential relevance to Aerial VLN.  Table~\ref{tab:datasets_unified} provides a unified overview of both categories.

\subsubsection{Dedicated Aerial VLN Datasets}

Dedicated datasets have evolved rapidly from simple instruction-trajectory pairs in minimal environments to large-scale, multi-modal corpora in photorealistic 3D reconstructions.

\paragraph{Early datasets}  LANI~\cite{misraMappingInstructionsActions2018} was the first dataset to pair natural language instructions with navigation trajectories from an aerial perspective, comprising 6,000 instruction sequences with first-person RGB observations and discrete actions. But LANI was constructed in a simplified simulation environment and supported only 2D navigation.

\paragraph{UAV-view datasets.}  Subsequent datasets introduced increasingly realistic 3D urban environments.
AerialVLN~\cite{liuAerialVLNVisionandlanguageNavigation2023} employs Unreal Engine(UE) to construct 25 virtual urban scenes with 8,400 trajectories and 4-DoF flight control, establishing the first widely adopted Aerial VLN benchmark.  CityNav~\cite{leeCityNavLargescaleDataset2025} constructs 3D environments covering parts of Cambridge and Birmingham based on the SensatUrban dataset~\cite{huSensatUrbanLearningSemantics2022}, providing 32,000 trajectories with natural language goal descriptions and human demonstration paths.  OpenFly~\cite{gaoOpenFlyComprehensivePlatform2025a} further extends environmental diversity by integrating scenes from UE, GTA-V and Google Earth, encompassing 100,000 trajectories across 18 real-world scenes. It is characterized by large-scale and fine-grained annotations supporting cross-city generalization evaluation.  OpenUAV~\cite{wangRealisticUAVVisionlanguage2024} advances the domain by transitioning from discrete 4-DoF actions to continuous 6-DoF flight trajectories across 22 scenes with 12,000 trajectories and multi-view cooperative perception.  
AirNav~\cite{cai2026airnavlargescalerealworlduav} employs real urban aerial data to construct 34 real-world scenes with 143,000 trajectories and 4-DoF flight control.
IndoorUAV~\cite{liuIndoorUAVBenchmarkingVisionlanguage2025} provides a rare indoor Aerial VLN dataset with over 5,000 high-fidelity trajectories paired with natural language instructions, supporting long-horizon navigation research in structured indoor environments. 

\paragraph{Satellite-view datasets}  The AVDN dataset~\cite{fanAerialVisionanddialogNavigation2023a} uses a satellite top-down perspective for dialog-based navigation, containing 3,000 multi-turn dialogs with human attention annotations for building localization.  
UAV-VLA~\cite{sautenkovUAVVLAVisionlanguageactionSystem2025} integrates satellite image processing with VLMs to generate path-action sets from 30 high-resolution satellite images.

\paragraph{Multi-view and cooperative datasets}  AeroDuo~\cite{wuAeroDuoAerialDuo2025} introduces a high-altitude/low-altitude dual-UAV cooperative paradigm with 13,000 image pairs for collaborative navigation.  EmbodiedCity~\cite{gaoEmbodiedCityBenchmarkPlatform2024} constructs a 3D digital twin of a section within Beijing's CBD, providing multi-view information from a single UAV and supporting dynamic traffic interaction.

\paragraph{Evaluation benchmarks}  Several recent datasets serve primarily as evaluation benchmarks rather than training datasets.  AeroVerse~\cite{yaoAeroVersereviewComprehensiveSurvey2025} presents a pipeline from virtual pre-training to real-world fine-tuning, defining five core evaluation: tasks spanning scene perception, spatial reasoning, navigation exploration, task planning and action decision-making.  
RefDrone~\cite{sunRefDroneChallengingBenchmark2025} and UAV-ON~\cite{xiaoUAVONBenchmarkOpenworld2025} provide goal-oriented benchmarks with multi-scale task suites emphasizing small-target localization.  
UAVBench~\cite{ferragUAVBenchOpenBenchmark2025} comprises 50,000 LLM-generated flight scenarios with structured multi-choice reasoning questions across 10 competency categories.  SpatialSky~\cite{zhangYourVLMSkyready2025} evaluates 13 fine-grained spatial intelligence capabilities with 1,000,000 samples.  
UrbanVideo-Bench~\cite{zhao-etal-2025-urbanvideo} assesses embodied cognitive capabilities across 16 tasks in four dimensions from aerial perspective.  
GeoText-1652~\cite{chu2025towards} evaluates cross-modal matching between natural language descriptions and drone-view imagery for geolocalization. 
CityCube~\cite{xuCityCubeBenchmarkingCrossview2026} integrates 18,100 images from 74 real-world cities and 2 virtual simulators to construct 5,022 meticulously annotated QA pairs for Aerial VLN evaluation.

\begin{table*}[t]
\centering
\caption{Unified overview of Aerial VLN datasets.  \textit{Dedicated} datasets pair flight trajectories with navigation instructions.  \textit{Underexploited} datasets contain UAV-perspective imagery with domain-specific annotations and potential for Aerial VLN applications.  DoF: degrees of freedom of the action space.  Modalities listed are those provided for each trajectory or image.}
\label{tab:datasets_unified}
\small
\begin{tabular}{l l c c c l}
\toprule
\textbf{Dataset} & \textbf{Category} & \textbf{Year} & \textbf{View} & \textbf{Scale} & \textbf{Key Characteristics} \\
\midrule
\multicolumn{6}{l}{\textit{Dedicated Aerial VLN Datasets}} \\
\addlinespace
LANI~\cite{misraMappingInstructionsActions2018} & Training & 2018 & UAV & 6K instr. & First aerial VLN corpus; 2D only \\
AerialVLN~\cite{liuAerialVLNVisionandlanguageNavigation2023} & Training+Eval & 2023 & UAV & 8.4K traj. & 25 UE scenes; 4-DoF discrete \\
CityNav~\cite{leeCityNavLargescaleDataset2025} & Training+Eval & 2024 & UAV & 32K traj. & Real point clouds; language-goal nav \\
AeroVerse~\cite{yaoAeroVerseUAVagentBenchmark2024} & Pre-train+Eval & 2024 & UAV & 10K--500K & Virtual pre-train to real fine-tune pipeline \\
EmbodiedCity~\cite{gaoEmbodiedCityBenchmarkPlatform2024} & Training+Eval & 2024 & UAV (multi) & -- & Beijing CBD digital twin; dynamic traffic \\
OpenFly~\cite{gaoOpenFlyComprehensivePlatform2025a} & Training+Eval & 2025 & UAV & 100K traj. & 18 real scenes; cross-city generalization \\
OpenUAV~\cite{wangRealisticUAVVisionlanguage2024} & Training+Eval & 2025 & UAV (multi) & 12K traj. & 22 scenes; continuous 6-DoF \\
IndoorUAV~\cite{liuIndoorUAVBenchmarkingVisionlanguage2025} & Training+Eval & 2025 & UAV & 5K+ traj. & Indoor; long-horizon VLN + VLA \\
AVDN~\cite{fanAerialVisionanddialogNavigation2023a} & Training+Eval & 2022 & Satellite & 3K dialogs & Multi-turn dialog; human attention maps \\
UAV-VLA~\cite{sautenkovUAVVLAVisionlanguageactionSystem2025} & Training & 2025 & Satellite & 30 images & Satellite-to-path-action generation \\
AeroDuo~\cite{wuAeroDuoAerialDuo2025} & Training+Eval & 2025 & Multi-alt. & 13K pairs & High/low altitude cooperative nav. \\
AirNav~\cite{cai2026airnavlargescalerealworlduav} & Training+Eval & 2026 & UAV & 143K traj. & Diverse/real urban aerial data and instructions \\
\addlinespace
\multicolumn{6}{l}{\textit{Evaluation Benchmarks}} \\
\addlinespace
RefDrone~\cite{sunRefDroneChallengingBenchmark2025} & Eval & 2025 & UAV & 8.5K img. & Fine-grained referring expression \\
UAV-ON~\cite{xiaoUAVONBenchmarkOpenworld2025} & Eval & 2025 & UAV & 1.3K targets & Open-world object-goal navigation \\
UAVBench~\cite{ferragUAVBenchOpenBenchmark2025} & Eval & 2025 & Structured & 50K scenarios & LLM-generated multi-choice reasoning \\
SpatialSky~\cite{zhangYourVLMSkyready2025} & Eval & 2025 & UAV+LiDAR & 1M samples & 13 spatial intelligence tasks \\
UrbanVideo-Bench~\cite{zhao-etal-2025-urbanvideo} & Eval & 2025 & UAV video & 1.5K videos & 16 embodied cognitive tasks \\
GeoText-1652~\cite{chu2025towards} & Eval & 2025 & Sat.+UAV & 276K pairs & Cross-modal geolocalization \\
CityCube~\cite{xuCityCubeBenchmarkingCrossview2026} & Eval & 2026 & Multi-view & 5K QA pairs & 5 cognitive dimensions \\
\addlinespace
\multicolumn{6}{l}{\textit{Underexploited Domain-Specific Datasets}} \\
\addlinespace
WEED-2C~\cite{tetilaRealtimeDetectionWeeds2024} & Detection & 2024 & UAV & 4K img. & Agriculture: weed species in soybean \\
InsPLAD~\cite{vieiraesilvaAttentionModulesImprove2024} & Detection & 2023 & UAV & 10.6K img. & Industry: power asset inspection \\
FloodNet~\cite{rahnemoonfarFloodNetHighResolution2021} & Segmentation & 2021 & UAV & 2.3K img. & Emergency: post-disaster assessment \\
VisDrone~\cite{zhuDetectionTrackingMeet2022} & Det.+Tracking & 2022 & UAV & 263 videos & Transport: daytime vehicle surveillance \\
TrafficNight~\cite{zhangTrafficNightAerialMultimodal2025} & Det.+Tracking & 2024 & UAV+IR & -- & Transport: nighttime + HD map fusion \\
MOCO~\cite{panMilitaryImageCaptioning2024} & Captioning & 2024 & UAV & 7.4K img. & Military: vehicle recognition \\
MMLA~\cite{klineMMLAMultienvironmentMultispecies2025} & Det.+Tracking & 2025 & UAV & 155K frames & Ecology: 6 wildlife species \\
RSVGD~\cite{zhanRSVGExploringData2023} & Grounding & 2022 & Satellite & 38K pairs & Remote sensing: language-guided localization \\
\bottomrule
\end{tabular}
\end{table*}

\subsubsection{Underexploited Domain-Specific Datasets}

Numerous datasets from vertical domains contain UAV-perspective imagery with semantic annotations and support Aerial VLN but have not yet been systematically integrated into VLN research.

In agriculture, the WEED-2C dataset~\cite{tetilaRealtimeDetectionWeeds2024} provides UAV images of soybean fields with species-level weed annotations.  
In industrial inspection, InsPLAD~\cite{vieiraesilvaAttentionModulesImprove2024} offers UAV images of electrical facilities with damage-level annotations across 17 asset classes.  
In emergency response, FloodNet~\cite{rahnemoonfarFloodNetHighResolution2021} provides UAV imagery with semantic annotations for damaged infrastructure, incorporating semantic segmentation and visual question answering components relevant to post-disaster assessment.  
In traffic surveillance, VisDrone~\cite{zhuDetectionTrackingMeet2022} and TrafficNight~\cite{zhangTrafficNightAerialMultimodal2025} provide daytime and nighttime UAV-captured traffic scenes respectively, with dense manual annotations for vehicle detection and tracking.  
In military reconnaissance, MOCO~\cite{panMilitaryImageCaptioning2024} focuses on battlefield environments from UAV perspectives with image captioning annotations for military vehicles.  
In remote sensing, RSVGD~\cite{zhanRSVGExploringData2023} is designed for visual grounding in remote sensing images, containing image-text pairs suitable for aerial spatial reasoning.  
In ecological monitoring, MMLA~\cite{klineMMLAMultienvironmentMultispecies2025} provides large-scale aerial wildlife images with textual annotations for species identification and behavior analysis.

The potential value of these datasets for Aerial VLN lies in their domain-specific and diverse visual content. Combined with instruction generation through LLMs or human annotation, underexploited Domain-Specific datasets enable the extension of Aerial VLN research in application domains beyond city navigation.

\subsubsection{Critical Gaps}

Aerial VLN related datasets are in infancy, with several limitations and gaps.
In dataset scale, Aerial VLN datasets are limited relative to ground VLN. Prevailing Aerial VLN datasets (e.g., OpenFly with 100,000 trajectories) remain an order of magnitude smaller than indoor VLN---a domain where datasets can reach millions of instruction-trajectory pairs~\cite{wang2023scaling}.
And the Aerial VLN environment lacks diversity, majority of dedicated datasets focus on urban street scenes. The underexploited datasets identified above could partially address this gap but require modifications to instruction-navigation matching. Additionally, the dialog-based UAV-view dataset is absent. The AVDN paradigm is currently supported by a single dataset~\cite{fanAerialVisionanddialogNavigation2023a} constructed from the satellite perspective.
In real-world data collection, nearly all dedicated Aerial VLN datasets are constructed in simulation. While simulators have reached high visual fidelity, they can't fully capture the noise, vision variability and dynamic conditions of real-world flight. UAV-Flow Colosseo~\cite{wangUAVflowColosseoRealworld2025} is among the few efforts to establish real-world evaluation, but the scale of real-world data remains negligible compared to simulated data. This gap is particularly consequential because the sim-to-real transfer problem (Section~\ref{sec:openproblems}) can't be meaningfully tackled without real-world benchmarks.

In standardization of evaluation, datasets differ in action space definitions (4-DoF versus 6-DoF, discrete versus continuous), success thresholds ($d_{\text{th}}$ ranging from 3\,m to 20\,m), and train/validation/test split conventions. These inconsistencies make cross-dataset performance comparison difficult and hinder the development of standardized benchmarks---a problem we quantify in Section~\ref{sec:analysis}.

\begin{figure*}[t]  
    \centering
    \includegraphics[width=1\linewidth]{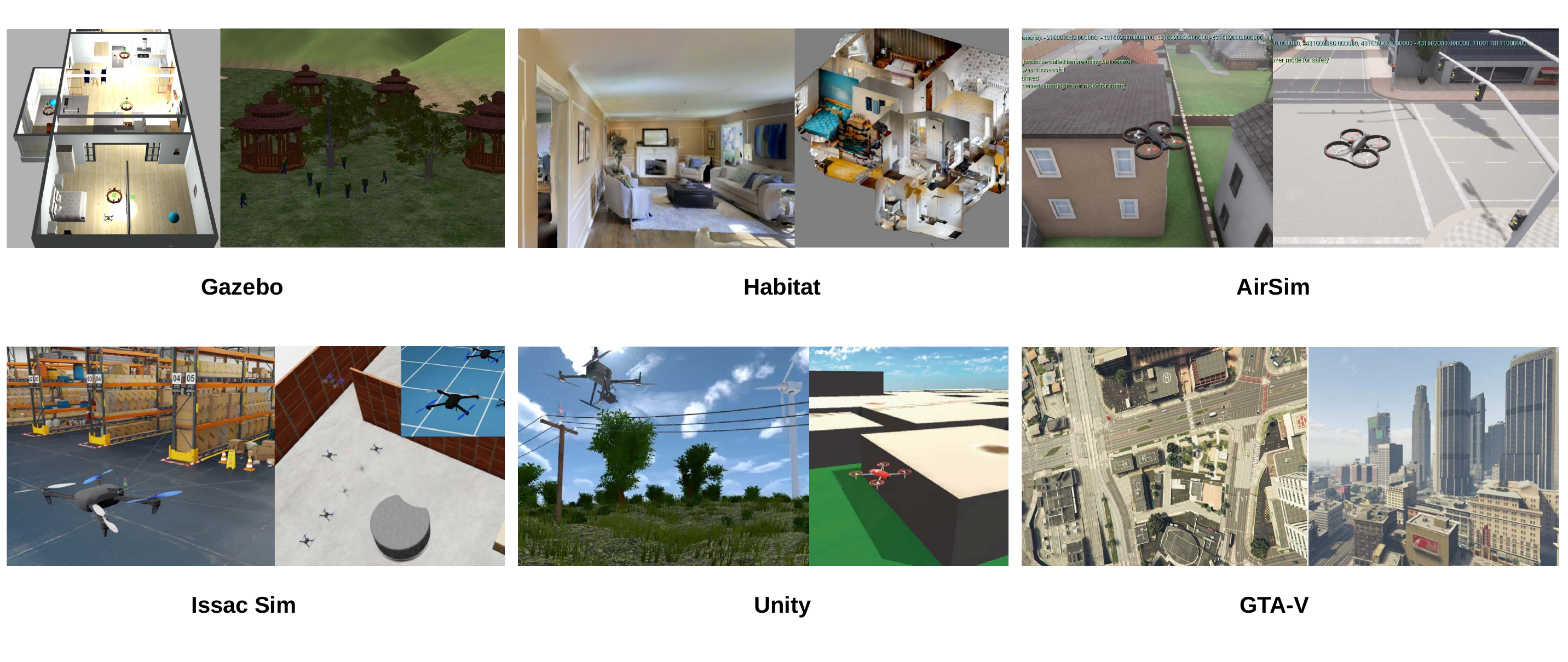}  
    \caption{Typical simulation platforms: Gazebo\cite{XiaGazebo1, XiaGazebo2}, Habitat\cite{changMatterport3DLearningRGBD2017a}, AirSim\cite{wangRealisticUAVVisionlanguage2024}, Issac Sim\cite{JacintoIssac, CuiIssac}, Unity\cite{SoaresUnity, WubbenUnity}, GTA-V\cite{JiGame4Loc, ji2025GTA}}
    \label{fig:Simulator}
\end{figure*}

\subsection{Simulation Platforms}\label{sec:simulators}

Simulation platforms provide configurable environments in which nearly all current Aerial VLN methods are developed, trained, and evaluated. Mainstream Aerial VLN platforms are surveyed in this subsection. Fig. \ref{fig:Simulator} illustrates typical simulation platforms.

\subsubsection{Platform Descriptions}

\paragraph{Gazebo} Gazebo~\cite{saxenaUAVVLNEndtoendVision2025} is a widely adopted open-source robotics simulator with a physics engine supporting rigid-body dynamics, joint constraints, and sensor characteristics.  Its integration with ROS makes it a natural choice for UAV algorithm and real-world deployment~\cite{XiaGazebo1, XiaGazebo2}. But the graphical rendering quality of Gazebo is relatively low, limiting its utility for vision-dependent tasks like VLN.

\paragraph{Habitat} Habitat~\cite{savva2019habitat} is designed specifically for embodied AI research, with core advantages in simulation speed and indoor scene datasets (Matterport3D~\cite{changMatterport3DLearningRGBD2017a}, Gibson~\cite{xiaGibsonEnvRealworld2018}).  It supports massively parallel simulation, providing the data throughput needed for VLN. But Habitat's applicability to Aerial VLN is fundamentally limited: its environments are predominantly indoor and physical interaction capabilities extend only to basic collision detection and movement constraints.

\paragraph{AirSim}  AirSim~\cite{shahAirSimHighfidelityVisual2018} is a high-fidelity simulation platform built on UE, targeting autonomous driving and UAV research. AirSim provides dedicated UAV flight dynamics models and user-friendly APIs that interface with mainstream deep learning frameworks.  It has been adopted as the primary simulator for several major Aerial VLN works~\cite{liuAerialVLNVisionandlanguageNavigation2023, wangRealisticUAVVisionlanguage2024}. But environment construction in AirSim relies on UE, requiring a certain technical foundation for user customization. And the process of importing custom models is relatively complex.

\paragraph{Isaac Sim}  Isaac Sim~\cite{nvidiaIsaacLabGPUaccelerated2025, yan2026aion} is NVIDIA's GPU-accelerated simulation platform built on the Omniverse framework, designed to unify high-quality visual rendering with high-precision physical simulation.  Its key strengths for Aerial VLN are GPU-native physics computation, built-in toolchains for large-scale synthetic data generation and native support for parallel RL. However, Isaac Sim remains in active early development, the algorithm toolchains and API interfaces are less mature than those of AirSim or Gazebo, the community of Aerial VLN users is small.

\paragraph{Unity} Unity is a widely used 3D simulation platform offering flexible scene construction, a built-in physics engine supporting flight dynamics and collision detection and sensor simulation plugins for LiDAR, RGB-D cameras, and IMU. Unity has been adopted for Aerial VLN methods that require custom environment construction with moderate engineering effort~\cite{zhangGroundedVisionlanguageNavigation2025, dominguez2026vln}.  Compared to AirSim, Unity offers greater flexibility in scene authoring but lacks AirSim's dedicated UAV dynamics models, requiring users to implement or integrate flight dynamics separately.

\paragraph{GTA-V}  GTA-V is a commercial 3D game engine that provides a richly detailed open-world environment with diverse terrain, dynamic traffic and pedestrian systems and a physics-based interaction engine.  These features make it attractive for constructing visually diverse and behaviorally realistic Aerial VLN scenarios.  Several VLN and geo-localization works have leveraged GTA-V environments~\cite{gaoOpenFlyComprehensivePlatform2025a, JiGame4Loc, YangOctopus}. However, GTA-V was not designed for robotics research and without native API for sensor simulation, flight dynamics or programmatic agent control. Researchers must rely on third-party modding tools to extract data and interface with learning frameworks.

\subsubsection{Comparative Assessment}

\begin{table*}[t]
\centering
\caption{Comparative assessment of simulation platforms for Aerial VLN.  Ratings ($\bullet\bullet\bullet$ = strong, $\bullet\bullet\circ$ = moderate, $\bullet\circ\circ$ = weak) reflect suitability for Aerial VLN specifically, not general robotics simulation.  ``Aerial VLN adoption'' indicates whether the platform has been used in published Aerial VLN works.}
\label{tab:simulators}
\small
\begin{tabular}{l c c c c c c}
\toprule
\textbf{Criterion} & \textbf{Gazebo} & \textbf{Habitat} & \textbf{AirSim} & \textbf{Isaac Sim} & \textbf{Unity} & \textbf{GTA-V} \\
\midrule
Visual realism & $\bullet\circ\circ$ & $\bullet\bullet\circ$ & $\bullet\bullet\bullet$ & $\bullet\bullet\bullet$ & $\bullet\bullet\circ$ & $\bullet\bullet\bullet$ \\
UAV flight dynamics & $\bullet\bullet\circ$ & $\circ\circ\circ$ & $\bullet\bullet\bullet$ & $\bullet\bullet\circ$ & $\bullet\circ\circ$ & $\circ\circ\circ$ \\
6-DoF continuous control & $\bullet\bullet\bullet$ & $\circ\circ\circ$ & $\bullet\bullet\bullet$ & $\bullet\bullet\bullet$ & $\bullet\bullet\circ$ & $\circ\circ\circ$ \\
Outdoor/city-scale env. & $\bullet\circ\circ$ & $\circ\circ\circ$ & $\bullet\bullet\bullet$ & $\bullet\bullet\circ$ & $\bullet\bullet\circ$ & $\bullet\bullet\bullet$ \\
Parallel training support & $\bullet\circ\circ$ & $\bullet\bullet\bullet$ & $\bullet\circ\circ$ & $\bullet\bullet\bullet$ & $\bullet\bullet\circ$ & $\circ\circ\circ$ \\
Ease of env.\ customization & $\bullet\bullet\circ$ & $\bullet\circ\circ$ & $\bullet\circ\circ$ & $\bullet\circ\circ$ & $\bullet\bullet\bullet$ & $\circ\circ\circ$ \\
Aerial VLN adoption & Moderate & None (aerial) & High & Emerging & Moderate & Moderate \\
\bottomrule
\end{tabular}
\end{table*}

Table~\ref{tab:simulators} summaries these simulation platforms in Aerial VLN. AirSim currently offers the best overall balance for Aerial VLN research, which combines high visual realism with dedicated UAV flight dynamics and has the largest adoption base in the published works. But the weak parallel training is insufficient for LLM/VLM-based methods. Isaac Sim addresses this scalability gap with GPU-native parallelism but lacks mature toolchains that AirSim provides. GTA-V offers unmatched environmental diversity and visual richness but provides no native support for the flight dynamics and sensor simulation that Aerial VLN requires, making it suitable primarily as a visual asset source rather than a complete simulation platform.  Habitat, despite its dominance in indoor VLN, is fundamentally unsuited for aerial tasks.

A critical limitation across all platforms is the absence of standardized Aerial VLN interfaces. Unlike indoor VLN, where the Habitat platform provides a unified API for environment loading, agent control and metric evaluation, Aerial VLN works adopt custom setting in different frameworks. Establishing a standardized simulation interface for Aerial VLN is a pressing infrastructure need that would accelerate progress across the field.

\subsection{Evaluation Metrics}\label{sec:metrics}
Evaluating Aerial VLN systems requires metrics that capture multiple dimensions of performance: whether the UAV reached its goal, how efficiently it navigated, how faithfully it followed the instruction, and---for LLM-centric methods---how reliably the cognitive components functioned.  We organize existing metrics into four categories by what they measure, Table~\ref{tab:metrics} provides a consolidated reference.

To formalize the evaluation metrics based on the Aerial VLN problem formulation (Section \ref{sec:formulation}), let $N$ denote the total number of evaluation episodes. For each episode $i \in \{1, \dots, N\}$, let $\mathbf{x}_{T,i}$ be the agent's final position, $\mathbf{x}_{g,i}$ be the target goal location, and $X_i = \{\mathbf{x}_{1,i}, \mathbf{x}_{2,i}, \dots, \mathbf{x}_{T_i,i}\}$ represent the executed trajectory of length $T_i$. Let $X^*_i = \{\mathbf{x}^*_{1,i}, \dots, \mathbf{x}^*_{T^*_i,i}\}$ denote the reference trajectory with shortest path length $l_i$, while the executed path length is $p_i = \sum_{t=1}^{T_i-1} \|\mathbf{x}_{t+1,i} - \mathbf{x}_{t,i}\|_2$.

\subsubsection{Navigation Success and Path Quality}

The most fundamental metric is the \textbf{success rate (SR)}, SR provides a binary pass/fail signal, the fraction of episodes in which the UAV's final position falls within the threshold distance $d_{\text{th}}$ of the goal (Eq. \ref{eq:success}).
\begin{equation}
    SR = \frac{1}{N} \sum_{i=1}^N \mathds{1}\left[\|\mathbf{x}_{T,i} - \mathbf{x}_{g,i}\|_2 \le d_{\text{th}}\right]
    \label{eq:sr}
\end{equation}

To capture path efficiency, \textbf{success weighted by path length (SPL)}~\cite{anderson2018vision} penalizes successful episodes in which the UAV took a substantially longer path than necessary, jointly measuring both success and navigation efficiency.
\begin{equation}
    SPL = \frac{1}{N} \sum_{i=1}^N \mathds{1}\left[\|\mathbf{x}_{T,i} - \mathbf{x}_{g,i}\|_2 \le d_{\text{th}}\right] \frac{l_i}{\max(p_i, l_i)}
    \label{eq:spl}
\end{equation}

\textbf{Normalized dynamic time warping (nDTW)} assesses trajectory similarity between the executed path and the reference path, evaluating whether the UAV navigated in a manner that approximates the intended route regardless of whether it ultimately reached the goal.
\begin{equation}
    nDTW = \frac{1}{N} \sum_{i=1}^N \exp\left(-\frac{\text{DTW}(X_i, X^*_i)}{D_{\max}}\right)
    \label{eq:ndtw}
\end{equation}
This is particularly informative for long-horizon aerial tasks where the UAV may follow most of the instruction correctly but stop slightly outside the success threshold.

Path deviation is quantified by \textbf{navigation error(NE)} metrics: \textbf{root mean square error (RMSE)} measures the deviation between actual and reference paths, and \textbf{mean absolute error (MAE)} measures the average per-step deviation from ground-truth positions.
\begin{equation}
    RMSE = \frac{1}{N} \sum_{i=1}^N \sqrt{\frac{1}{T_i} \sum_{t=1}^{T_i} \|\mathbf{x}_{t,i} - \mathbf{x}^*_{t,i}\|_2^2}
    \label{eq:rmse}
\end{equation}
\begin{equation}
    MAE = \frac{1}{N} \sum_{i=1}^N \frac{1}{T_i} \sum_{t=1}^{T_i} \|\mathbf{x}_{t,i} - \mathbf{x}^*_{t,i}\|_2
    \label{eq:mae}
\end{equation}

\subsubsection{LLM Reasoning Fidelity}

As LLMs assume the role of cognitive core in Aerial VLN systems, metrics are needed to evaluate the quality of the LLM's reasoning and instruction processing, independent of the downstream navigation outcome~\cite{changSurveyEvaluationLarge2024}.

The \textbf{instruction completion rate (ICR)} measures execution fidelity within the navigation process either the overall task completion or the count of specified sub-instructions that were successfully executed. Assuming the instruction $\mathcal{L}_i$ is decomposed into $K_i$ sub-instructions,
\begin{equation}
    ICR = \frac{1}{N} \sum_{i=1}^N \frac{1}{K_i} \sum_{k=1}^{K_i} \mathds{1}\left[\text{sub-instruction } k \text{ is completed}\right]
    \label{eq:icr}
\end{equation}
ICR is more informative than SR for methods that decompose instructions into subgoals.

The \textbf{zero-shot generalization success rate (ZGSR)} evaluates performance of pre-trained models across novel tasks and scenes without fine-tuning. Given an unseen evaluation dataset $\mathcal{D}_{\text{unseen}}$ with $N_{\text{unseen}}$ episodes,
\begin{equation}
    ZGSR = \frac{1}{N_{\text{unseen}}} \sum_{j \in \mathcal{D}_{\text{unseen}}} \mathds{1}\left[\|\mathbf{x}_{T,j} - \mathbf{x}_{g,j}\|_2 \le d_{\text{th}}\right]
    \label{eq:zgsr}
\end{equation}
ZGSR is increasingly important as the field shifts toward LLM-based methods that claim generalization as a core advantage.

\textbf{Output confidence}, represented by probability distributions over the model's possible outputs, gauges the LLM's self-assessed certainty in what generated.
\begin{equation}
    \text{Confidence}_t = \max_{a \in \mathcal{A}} P(a_t = a \mid \mathcal{L}, o_{1:t})
    \label{eq:confidence}
\end{equation}

\textbf{Cross-modal alignment score (CAS)} provides a metric for evaluating the semantic consistency between natural language instructions and the visual information perceived during navigation.
\begin{equation}
    CAS = \frac{1}{T} \sum_{t=1}^T \text{sim}(f_{\mathcal{L}}(\mathcal{L}), f_{\mathcal{V}}(\mathcal{V}_t))
    \label{eq:cas}
\end{equation}
where $f_{\mathcal{L}}$ and $f_{\mathcal{V}}$ extract features from the instruction and visual observation, respectively, and $\text{sim}(\cdot, \cdot)$ computes their similarity. Both metrics remain underutilized in current Aerial VLN evaluations, partly because they require access to model internals that are not always available for proprietary LLMs.

\subsubsection{Safety and Operational Efficiency} 

The \textbf{flight energy efficiency (FEE)} assesses battery utilization during flight, which is critical for real-world missions where energy constraints directly limit operational range. The \textbf{security violation rate (SVR)} quantifies the frequency of unsafe incidents such as collisions with obstacles, airspace boundary violations, or proximity infringements. Let $C(s_t)$ indicate a safety violation at state $s_t$,
\begin{equation}
    FEE = \frac{1}{N} \sum_{i=1}^N \frac{p_i}{\int_0^{T_i} P_{i}(t) dt}
    \label{eq:fee}
\end{equation}

\begin{equation}
    SVR = \frac{1}{N} \sum_{i=1}^N \mathds{1}\left[\sum_{t=1}^{T_i} C(s_{t,i}) > 0\right]
    \label{eq:svr}
\end{equation}
where $P_{i}(t)$ is the power consumption at time $t$. FEE and SVR are rarely reported in current Aerial VLN evaluations because most methods are developed and evaluated in simulators where energy is unconstrained and collisions have no physical consequence. As Aerial VLN methods mature toward real-world applications, safety and efficiency metrics will need to become standard components of the evaluation protocol.

\subsubsection{Summary and Metric Gaps}

\begin{table*}[t]
\centering
\caption{Taxonomy of evaluation metrics for Aerial VLN.  Metrics are grouped by the dimension of performance they capture.  ``Adoption'' indicates how widely each metric is reported in current Aerial VLN publications.}
\label{tab:metrics}
\small
\begin{tabular}{l l l}
\toprule
\textbf{Category} & \textbf{Metric} & \textbf{Adoption} \\
\midrule
\multirow{5}{*}{\textit{Nav.\ Success \& Path}} & SR (Success Rate) & Universal \\
 & SPL (Success $\times$ Path Length) & Common \\
 & nDTW (Norm.\ Dyn.\ Time Warp) & Moderate \\
 & RMSE / MAE (Path Deviation) & Moderate \\
\addlinespace
\multirow{4}{*}{\textit{LLM Reasoning}} & ICR (Instruction Completion) & Rare \\
 & ZGSR (Zero-shot Generalization) & Rare \\
 & Output Confidence & Very rare \\
 & CAS (Cross-modal Alignment) & Very rare \\
\addlinespace
\multirow{2}{*}{\textit{Safety \& Efficiency}} & FEE (Flight Energy Efficiency) & Very rare \\
 & SVR (Security Violation Rate) & Very rare \\
\bottomrule
\end{tabular}
\end{table*}

Table~\ref{tab:metrics} reveals a stark adoption imbalance: SR and SPL are reported near-universally, while LLM reasoning metrics and safety/efficiency metrics appear only sporadically.  This imbalance means that the current Aerial VLN works primarily evaluate whether the UAV reached its goal, with limited insight into how it reasoning, how safely and efficiently it flying. Beyond this adoption gap, some specific metric gaps deserve attention. 
Many LLM-centric methods (Sections~\ref{sec:e2e} and~\ref{sec:hierarchical}) decompose lengthy instructions into subgoals, but there's no metric for instruction decomposition quality to evaluate whether this decomposition is correct, complete, or appropriately granular. 
In real-time responsiveness, none of the established metrics capture inference latency or planning frequency.
For AVDN methods (Section~\ref{sec:dialog}), existing metrics evaluate only the final navigation outcome, not the quality of the dialog interaction.
Viewpoint variation is a defining challenge of Aerial VLN (Section~\ref{sec:context}), but there's no metric specifically quantifies a method's robustness to altitude and orientation changes.  
In cross-benchmark comparability, even for universally adopted metrics like SR, direct comparison across benchmarks is complicated by differing success thresholds ($d_{\text{th}}$ ranges from 3\,m to 20\,m), action spaces (discrete versus continuous), and episode conditions. This comparability problem constrains the cross-method analysis in Section~\ref{sec:analysis} and underscores the need for standardized evaluation protocols discussed in Section~\ref{sec:openproblems}.

\section{Comparative Analysis and Discussion}\label{sec:analysis}
Having surveyed the landscape of Aerial VLN across methodologies and evaluation infrastructure, this section synthesizes the accumulated empirical evidence into a unified comparative analysis and discussion, which are structured along three dimensions: a quantitative comparison across multiple benchmarks tracing performance trends over successive generations of methods; an analysis of fundamental architectural trade-offs in action space, policy hierarchy, and training paradigm; and an examination of the persistent simulation-to-reality gap and emerging strategies to bridge it.

\subsection{Quantitative Comparison Across Benchmarks}
Most existing Aerial VLN benchmarks have only recently been introduced and have yet to achieve the widespread adoption necessary for comprehensive quantitative evaluation. Moreover, certain datasets are designed for highly specific tasks, which inherently limits their generalizability. We therefore select four relatively mature benchmarks that have been evaluated across multiple methodologies for systematic comparison: AerialVLN-S, AVDN, OpenUAV, and CityNav. A chronological examination of performance metrics reveals a clear paradigm shift within the field. Early ground-based Seq2Seq and cross-modal attention (CMA) architectures are ill-suited to the 6-DoF dynamics, altitude-dependent viewpoint shifts, and long-horizon planning demands of aerial navigation, typically yielding success rates (SR) below 10\% in unseen environments. By contrast, recent frameworks that leverage large language models (LLMs) or vision-language models (VLMs), explicit spatial reasoning, and dynamic physical control mechanisms demonstrate substantial improvements in navigation error (NE) and success weighted by path length (SPL).

\subsubsection{AerialVLN}
AerialVLN-S is the small-scene variant of the AerialVLN benchmark. As shown in Table \ref{tab:aerialvln_s_comparison}, conventional baselines such as Seq2Seq and CMA demonstrate limited navigation efficacy, with Validation Unseen SRs of only 2.3\% and 3.2\%, respectively. Incorporating structured spatial representations yields measurable gains: the Grid-based View method achieves an SR of 20.8\% on the Validation Seen split. However, the CityNavAgent framework demonstrates superior generalization on the Validation Unseen split, attaining an SR of 11.7\% and reducing NE to 60.2 m. This highlights the critical role of advanced spatial memory and topological mapping in preventing catastrophic navigation failures in large-scale aerial environments.

\begin{table*}[htbp]
\centering
\caption{Performance comparison of selected methods on the AerialVLN-S dataset.}
\label{tab:aerialvln_s_comparison}
\resizebox{\textwidth}{!}{
\begin{tabular}{lcccccccc}
\toprule
\multirow{2}{*}{\textbf{Method}} & \multicolumn{4}{c}{\textbf{Validation Seen}} & \multicolumn{4}{c}{\textbf{Validation Unseen}} \\
\cmidrule(lr){2-5} \cmidrule(lr){6-9}
 & SR (\%) $\uparrow$ & OSR (\%) $\uparrow$ & SDTW (\%) $\uparrow$ & NE (m) $\downarrow$ & SR (\%) $\uparrow$ & OSR (\%) $\uparrow$ & SDTW (\%) $\uparrow$ & NE (m) $\downarrow$ \\
\midrule
Random~\cite{liuAerialVLNVisionandlanguageNavigation2023} & 0.0 & 0.0 & 0.0 & 109.6 & 0.0 & 0.0 & 0.0 & 149.7 \\
Seq2Seq~\cite{liuAerialVLNVisionandlanguageNavigation2023} & 4.8 & 19.8 & 1.6 & 146.0 & 2.3 & 11.7 & 0.7 & 218.9 \\
CMA~\cite{liuAerialVLNVisionandlanguageNavigation2023} & 3.0 & 23.2 & 0.6 & 121.0 & 3.2 & 16.0 & 1.1 & 172.1 \\
LAG~\cite{liuAerialVLNVisionandlanguageNavigation2023} & 7.2 & 15.7 & 2.4 & 90.2 & 5.1 & 10.5 & 1.4 & 127.9 \\
Grid-based View~\cite{zhaoAerialVisionandlanguageNavigation2025} & \textbf{20.8} & 33.4 & \textbf{10.2} & \textbf{70.3} & 7.4 & 16.1 & 2.5 & 121.3 \\
STMR~\cite{gaoAerialVisionandlanguageNavigation2024,gaoExploringSpatialRepresentation2025} & 12.6 & 31.6 & 2.7 & 96.3 & 10.8 & 23.0 & - & 119.5 \\
CityNavAgent~\cite{zhangCityNavAgentAerialVisionandlanguage2025} & 13.9 & 30.2 & 5.1 & 80.8 & \textbf{11.7} & \textbf{35.2} & \textbf{5.0} & \textbf{60.2} \\
OpenFly\cite{xuAerialVisionLanguageNavigation2025} & 8.1 & 21.8 & 1.6 & 127.2 & 7.6 & 18.2 & 1.5 & 113.8 \\
Unified Aerial VLN\cite{xuAerialVisionLanguageNavigation2025} & 11.4 & \textbf{37.7} & 6.3 & 79.6 & 8.1 & 28.9 & 2.2 & 95.8 \\
\bottomrule
\end{tabular}
}
\end{table*}

\subsubsection{AVDN}
Table \ref{tab:avdn_comparison} summarizes performance on the AVDN benchmark, which uniquely incorporates dialog history and evaluates Goal Progress (GP). History-aware baselines such as HAA-LSTM establish a foundational Test Unseen SR of 14.1\%, while more recent visual-grounding architectures such as VAG push the Validation Unseen SPL to 22.1\%. A particularly notable advancement is the SkyVLN framework, which tightly couples high-level VLN planning with low-level predictive physical control via nonlinear model predictive control (NMPC). This integration achieves a Test Unseen SPL of 28.1\% and SR of 42.4\%, representing a substantial absolute improvement over purely perceptual approaches. These results underscore that combining visual-linguistic alignment with robust kinematic control is essential for reliable navigation under dialog guidance.

\begin{table*}[htbp]
\centering
\caption{Performance comparison of selected methods on the AVDN dataset.}
\label{tab:avdn_comparison}
\resizebox{\textwidth}{!}{
\begin{tabular}{lccccccccc}
\toprule
\multirow{2}{*}{\textbf{Method}} & \multicolumn{3}{c}{\textbf{Validation Seen}} & \multicolumn{3}{c}{\textbf{Validation Unseen}} & \multicolumn{3}{c}{\textbf{Test Unseen}} \\
\cmidrule(lr){2-4} \cmidrule(lr){5-7} \cmidrule(lr){8-10}
 & SPL (\%) $\uparrow$ & SR (\%) $\uparrow$ & GP (\%) $\uparrow$ & SPL (\%) $\uparrow$ & SR (\%) $\uparrow$ & GP (\%) $\uparrow$ & SPL (\%) $\uparrow$ & SR (\%) $\uparrow$ & GP (\%) $\uparrow$ \\
\midrule
Random~\cite{fanAerialVisionanddialogNavigation2023a} & 0.5 & 1.6 & -84.1 & 0.2 & 1.0 & -81.4 & 0.5 & 1.1 & -86.6 \\
HAA-LSTM~\cite{fanAerialVisionanddialogNavigation2023a} & 11.6 & 13.0 & 50.3 & 18.3 & 20.0 & 54.4 & 12.6 & 14.1 & 50.8 \\
HAA-Transformer~\cite{fanAerialVisionanddialogNavigation2023a} & 14.7 & 17.3 & 56.3 & 16.5 & 20.4 & 55.2 & 12.9 & 15.7 & 54.2 \\
TG-GAT~\cite{suTargetgroundedGraphawareTransformer2023} & 12.9 & 16.0 & 56.9 & 18.8 & 23.3 & 54.3 & 15.1 & 18.7 & 56.5 \\
DBDP~\cite{wangDualbranchDynamicPerception2025} & 15.6 & 18.3 & 58.7 & 18.6 & 22.6 & 55.8 & 13.7 & 16.3 & 55.8 \\
FELA\cite{suLearningFinegrainedAlignment} & 15.3 & 18.8 & \textbf{60.7} & 19.2 & 23.9 & \textbf{64.1} & 17.6 & 21.9 & \textbf{61.4} \\
VAG~\cite{qiaoEnhancingVisualAligning2025} & \textbf{19.0} & \textbf{21.9} & 57.6 & \textbf{22.1} & \textbf{24.6} & 59.5 & 20.6 & 22.9 & 59.6 \\
SkyVLN~\cite{liSkyVLNVisionandlanguageNavigation2025} & 14.7 & 17.3 & - & 16.6 & 20.4 & - & \textbf{28.1} & \textbf{42.4} & - \\
\bottomrule
\end{tabular}
}
\end{table*}

\subsubsection{OpenUAV}
The OpenUAV benchmark (Table \ref{tab:openuav_comparison}) evaluates agents in continuous, physically realistic simulation environments, where sample efficiency and multi-agent collaboration emerge as decisive factors. On the Test Seen split, OpenVLN achieves a 14.39\% SR using only 25\% of the available training data, demonstrating strong data efficiency. In the more demanding Unseen Map split, single-UAV approaches such as TravelUAV suffer significant performance degradation. In contrast, the AeroDuo model introduces a collaborative dual-UAV architecture that achieves a 16.57\% SR and reduces NE to 84.31 m. This superiority indicates that multi-UAV collaborative perception effectively compensates for the limited field-of-view inherent to single-agent aerial platforms.

\begin{table*}[htbp]
\centering
\caption{Performance comparison of selected methods on the OpenUAV benchmark.}
\label{tab:openuav_comparison}
\resizebox{0.7\textwidth}{!}{
\begin{tabular}{llcccc}
\toprule
\multirow{2}{*}{\textbf{Benchmark}} & \multirow{2}{*}{\textbf{Method}} & \multicolumn{4}{c}{\textbf{Metrics}} \\
\cmidrule(lr){3-6}
 & & NE (m) $\downarrow$ & SR (\%) $\uparrow$ & OSR (\%) $\uparrow$ & SPL (\%) $\uparrow$ \\
\midrule
\multirow{4}{*}{\shortstack{OpenUAV\\Test Seen}} 
 & Random & 222.20 & 0.14 & 0.21 & 0.07 \\
 & CMA (100\%)~\cite{anderson2018vision} & 135.73 & 8.37 & 18.72 & 7.90 \\
 & TravelUAV (25\%)~\cite{wangRealisticUAVVisionlanguage2024} & 132.59 & 11.59 & 24.50 & 10.45 \\
 & OpenVLN (25\%)~\cite{linOpenVLNOpenworldAerial2025} & \textbf{125.97} & \textbf{14.39} & \textbf{28.03} & \textbf{12.94} \\
\midrule
\multirow{4}{*}{\shortstack{OpenUAV\\Unseen Map}} 
 & Random & 199.42 & 0.00 & 0.00 & 0.00 \\
 & CMA~\cite{anderson2018vision} & 166.31 & 0.00 & 0.57 & 0.00 \\
 & TravelUAV (L1)~\cite{wangRealisticUAVVisionlanguage2024} & 107.91 & 6.86 & 17.14 & 5.89 \\
 & AeroDuo~\cite{wuAeroDuoAerialDuo2025} & \textbf{84.31} & \textbf{16.57} & \textbf{28.57} & \textbf{13.86} \\
\bottomrule
\end{tabular}
}
\end{table*}

\subsubsection{CityNav}
Table \ref{tab:citynav_comparison} presents a comparative analysis on CityNav, a large-scale real-world 3D dataset partitioned by Validation/Test and Seen/Unseen difficulty levels. We note that GeoNav employs a custom difficulty-based taxonomy that does not explicitly align with CityNav's official splits; both categorization schemes are therefore denoted within the table for transparency. Traditional architectures fail to scale to real-world complexity, with SRs stagnating below 5\% on Medium and Hard splits. The introduction of multimodal large language models (MLLMs) produces a significant performance bifurcation. GeoNav, which relies on explicit geospatial querying, excels on Easy and Medium environments with SRs of 26.53\% and 22.92\%, respectively. However, on the Test Unseen (Hard) split, SA-GCS—leveraging semantic-aware curriculum scheduling—demonstrates superior robustness, achieving an SR of 24.55\% and an SPL of 22.86\%. This contrast suggests that while explicit spatial reasoning is highly effective in moderately familiar topologies, curriculum-driven reinforcement learning (RL) confers greater adaptability in complex, entirely novel urban landscapes.

\begin{table*}[htbp]
\centering
\caption{Performance comparison of selected methods on the CityNav benchmark.}
\label{tab:citynav_comparison}
\resizebox{\textwidth}{!}{
\begin{tabular}{lcccccccccccc}
\toprule
\multirow{2}{*}{\textbf{Method}l} & \multicolumn{4}{c}{\textbf{Validation Seen (Easy)}} & \multicolumn{4}{c}{\textbf{Validation Unseen (Medium)}} & \multicolumn{4}{c}{\textbf{Test Unseen (Hard)}} \\
\cmidrule(lr){2-5} \cmidrule(lr){6-9} \cmidrule(lr){10-13}
 & NE (m) $\downarrow$ & SR (\%) $\uparrow$ & OSR (\%) $\uparrow$ & SPL (\%) $\uparrow$ & NE (m) $\downarrow$ & SR (\%) $\uparrow$ & OSR (\%) $\uparrow$ & SPL (\%) $\uparrow$ & NE (m) $\downarrow$ & SR (\%) $\uparrow$ & OSR (\%) $\uparrow$ & SPL (\%) $\uparrow$ \\
\midrule
Seq2Seq~\cite{leeCityNavLargescaleDataset2025} & 58.50 & 8.43 & 17.31 & 7.28 & 78.60 & 5.13 & 10.90 & 4.65 & 98.10 & 3.81 & 13.92 & 2.79 \\
CMA~\cite{leeCityNavLargescaleDataset2025} & 68.00 & 6.25 & 13.28 & 5.40 & 75.90 & 4.38 & 9.29 & 3.90 & 94.60 & 4.68 & 12.01 & 4.05 \\
AerialVLN~\cite{leeCityNavLargescaleDataset2025} & \textbf{56.60} & 10.16 & 22.20 & 7.89 & 72.70 & 6.35 & 15.24 & 5.06 & 85.10 & 6.72 & 18.21 & 5.16 \\
FlightGPT~\cite{caiFlightGPTGeneralizableInterpretable2025} & 66.10 & 17.57 & 30.26 & 15.78 & 68.10 & 14.69 & 29.33 & 13.24 & 76.20 & 21.20 & 35.38 & 19.24 \\
SA-GCS~\cite{caiFlightGPTGeneralizableInterpretable2025} & 59.68 & 18.69 & 31.98 & \textbf{17.26} & 63.76 & 16.88 & 31.48 & 15.89 & \textbf{68.42} & \textbf{24.55} & \textbf{37.36} & \textbf{22.86} \\
\midrule
GeoNav~\cite{xuGeoNavEmpoweringMLLMs2025} & 59.86 & \textbf{26.53} & \textbf{73.47} & 12.05 & \textbf{53.80} & \textbf{22.92} & \textbf{39.58} & \textbf{17.06} & 68.90 & 16.67 & 22.92 & 12.49 \\
\bottomrule
\end{tabular}
}
\end{table*}

In summary, the cross-benchmark analysis points to several multifaceted directions for the future of Aerial VLN. The collective empirical evidence underscores that overcoming the unique challenges of aerial navigation requires a paradigm shift from purely perceptual cross-modal matching toward physically grounded, embodied intelligence. As demonstrated by the AVDN and AerialVLN-S evaluations, integrating high-level linguistic reasoning with low-level kinematic control and persistent topological memory is indispensable for mitigating trajectory drift in continuous 3D spaces. The OpenUAV results further highlight a critical structural evolution: multi-agent collaborative perception can break the sensory bottleneck imposed by single-viewpoint UAVs. Scaling to real-world, city-level complexity additionally demands sophisticated adaptation strategies, as the CityNav results demonstrate that explicit geospatial reasoning and curriculum-driven RL are complementary rather than competing approaches. Numerous real-world experiments \cite{chenGRADNAVVisionlanguageModel2026, songSoraNavAdaptiveUAV2025, cai2026airnavlargescalerealworlduav} have further confirmed that integrating LLMs into Aerial VLN represents a significant breakthrough for aerial embodied intelligence. Ultimately, the next generation of Aerial VLN systems will be defined not merely by linguistic comprehension, but by their capacity for spatial cognition, multi-platform synergy, and robust physical execution in open-world environments.


\subsection{Architectural Trade-offs}
The advancement of Aerial VLN involves fundamental trade-offs across action space design, policy architecture, and training paradigm. Each design choice carries distinct advantages and limitations with respect to computational efficiency, navigational accuracy, and open-world generalizability.

\subsubsection{Discrete vs. Continuous Action Spaces}
Discrete action spaces simplify policy learning and integrate naturally with traditional sequence models, offering high computational efficiency. However, they fail to capture the complex 6-DoF dynamics of real UAVs and frequently lead to trajectory drift. Continuous action spaces, by contrast, produce smooth, aerodynamically feasible, and collision-free trajectories, significantly enhancing physical realism and safety. This comes at the cost of substantially greater optimization difficulty, cross-modal alignment complexity, and real-time inference latency.

\subsubsection{End-to-End, Hierarchical, and Multi-agent Policy}
End-to-end policies map perceptual inputs directly to control commands, benefiting from low inference latency. However, early Seq2Seq and attention-based approaches suffer from poor generalization, and while LLM-based end-to-end systems show stronger performance, severe trajectory drift and navigation forgetting persist during long-horizon tasks. Hierarchical policies mitigate these issues by decoupling high-level semantic planning from low-level execution, enabling robust reasoning and precise obstacle avoidance—but at the cost of cascading errors and module synchronization overhead. Multi-agent policies transcend the perceptual limitations of single agents through collective decision-making, though this requires managing inter-agent communication, data synchronization, and cross-view feature alignment.

\subsubsection{Fine-tuned vs. Zero-shot Training}
Domain-specific fine-tuning achieves high accuracy and tight visual-linguistic grounding within known environments, but is prone to overfitting and depends heavily on expensive human-annotated trajectories, leading to sharp performance degradation in unseen splits. Zero-shot paradigms that leverage frozen pre-trained LLMs demonstrate strong open-world generalization and commonsense reasoning without task-specific data. However, absent domain adaptation, zero-shot models frequently struggle with precise visual grounding, particularly when mapping nuanced instructions to the unfamiliar top-down visual perspectives unique to aerial navigation.

\subsection{The Simulation-Reality Gap}
Bridging the simulation-to-reality (sim-to-real) divide remains one of the most pressing challenges in Aerial VLN. Recent methods validated through real-world physical deployments have pursued three distinct technological trajectories.

The first is the construction of high-fidelity simulation environments. Frameworks such as GRAD-Nav++ \cite{chenGRADNAVVisionlanguageModel2026} and SINGER \cite{adangSINGEROnboardGeneralist2025} have pioneered the use of 3D Gaussian Splatting (3DGS) to build photorealistic, language-embedded flight simulators. GRAD-Nav++ achieves a 67\% SR on trained tasks and 50\% SR on entirely unseen tasks in real-world deployment. SoraNav \cite{songSoraNavAdaptiveUAV2025} further addresses dynamic discrepancies between physical UAVs and virtual environments through a hardware-software digital twin platform.

The second trajectory is the emergence of zero-shot, training-free inference. Architectures such as SPF \cite{huSeePointFly2025} and STMR \cite{gaoAerialVisionandlanguageNavigation2024} leverage frozen pre-trained VLMs, reformulating navigation decisions as explicit 2D spatial grounding or topological reasoning problems. This training-free paradigm demonstrates robust real-world generalizability without requiring environment-specific adaptation.

The third trajectory involves the direct use of real-world perception data and flight trajectories for training and evaluation. UAV-Flow Colosseo \cite{wangUAVflowColosseoRealworld2025} comprises over 30,000 real-world episodes and more than 100 hours of teleoperated flight recordings across diverse large-scale campuses. AirNav \cite{cai2026airnavlargescalerealworlduav} provides over 143,000 real-world navigation episodes derived from urban aerial imagery, with diverse natural-language instructions spanning two large-scale cities.

Despite these advances, the vast majority of current Aerial VLN methods remain confined to simulated environments for training, validation, and testing. Real-world physical deployments are still rare and often yield suboptimal performance. Closing the sim-to-real gap remains a fundamental and urgent open challenge for the field.

\begin{figure*}[t]  
    \centering
    \includegraphics[width=0.9\linewidth]{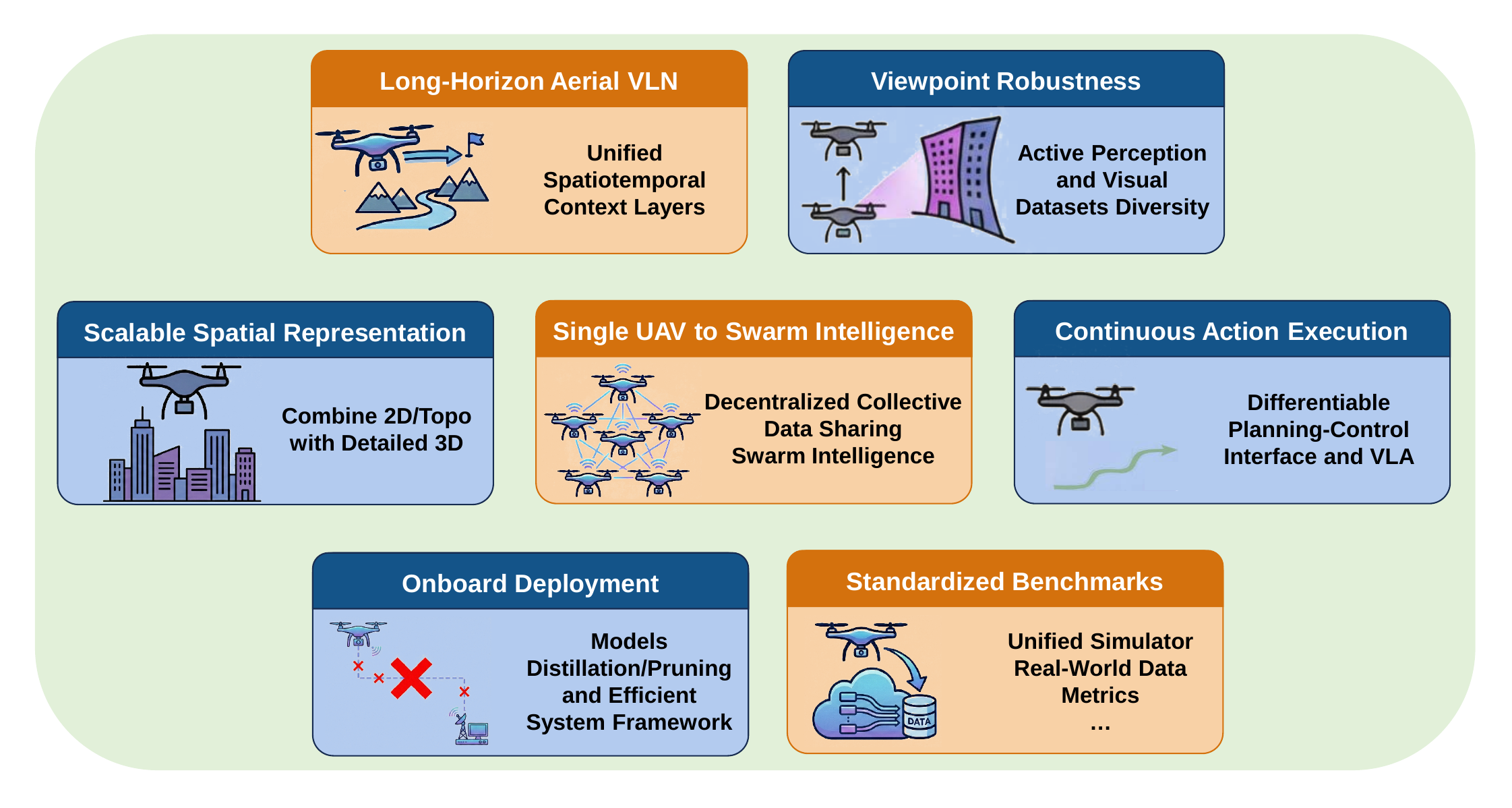}  
    \caption{Open problems of Aerial VLN.}
    \label{fig:Open_Problem}
\end{figure*}

\section{Open Problems and Research Directions}\label{sec:openproblems}

The comparative analysis in Section~\ref{sec:analysis} and the infrastructure assessment in Section~\ref{sec:infrastructure} reveal that despite rapid methodological progress, Aerial VLN faces several fundamental open problems that constrain both performance and deployability. This section synthesizes these problems into seven thematic areas, as illustated in Fig. \ref{fig:Open_Problem}.  For each, we define the problem with reference to evidence from earlier sections, review current approaches and limitations, then propose specific research directions.

\subsection{Long-Horizon Navigation and Instruction Grounding}
\label{sec:op_longhorizon}

\textbf{Problem.}  Aerial VLN instructions are substantially longer and more structurally complex than their indoor counterparts (Section~\ref{sec:context}): they interleave horizontal navigation, vertical maneuvers, temporal sequencing, and conditional logic over trajectories spanning hundreds of meters.  Maintaining coherent alignment between the UAV's cumulative trajectory and the full instruction over extended episodes is a central challenge, requiring the agent to track progress, remember past landmarks, and anticipate future segments simultaneously.

\textbf{Current approaches and limitations.}  Two decomposition strategies dominate.  Rule-based or pattern-based approaches use pre-trained language models such as BERT~\cite{zhaoAerialVisionandlanguageNavigation2025} or RoBERTa~\cite{qiaoEnhancingVisualAligning2025} to segment instructions into sub-instructions based on syntactic cues.  LLM-based generative approaches~\cite{liSkyVLNVisionandlanguageNavigation2025, zhangCityNavAgentAerialVisionandlanguage2025, xuGeoNavEmpoweringMLLMs2025, zhangMem2EgoEmpoweringVisionlanguage2025} leverage the contextual reasoning capabilities of LLMs to decompose instructions into semantically coherent subgoals. 
Both paradigms exhibit limitations in decomposition accuracy and reasoning of 3D spatial relationships.  Rule-based methods rely on surface-level syntactic patterns that miss the causal and spatial dependencies between instruction segments.  LLM-based decomposition produces more natural subgoals but is prone to hallucination---generating subgoals that are physically implausible or arbitrarily concocted  Furthermore, decomposed subgoals must be dynamically aligned with real-time perception in changing environments, and this alignment degrades as the horizon extends.

\textbf{Research directions.} Three directions are promising. The first is unified spatiotemporal context models~\cite{jiangLongFlyLonghorizonUAV2025} that maintain a persistent representation integrating spatial and temporal memory with instruction state.  The second is embodied world models~\cite{hafnerMasteringDiverseControl2025, bar2025navigation} that support predictive reasoning for future trajectory segments and evaluate subgoal feasibility.  The third is hierarchical instruction representations that preserve the full instruction structure while exposing different fine-grained sub-instructions to different navigation components.

\subsection{Viewpoint Robustness and Cross-View Alignment}
\label{sec:op_viewpoint}

\textbf{Problem.}  The continuous and severe viewpoint changes caused by UAV flight cause the same landmark to produce nonlinearly distorted visual representations across observation steps (Section~\ref{sec:context}).  This disrupts the stable visual-semantic associations that cross-modal alignment methods depend on, leading to failures in landmark recognition, instruction grounding and progress tracking.

\textbf{Current approaches and limitations.}  Two mitigation strategies have been explored. One employs Aerial VLN with open-vocabulary detection models such as Grounding DINO~\cite{gaoAerialVisionandlanguageNavigation2024, liSkyVLNVisionandlanguageNavigation2025}, GLIP~\cite{liuNavAgentMultiscaleUrban2024}, or VLMs like GPT-4V~\cite{zhangCityNavAgentAerialVisionandlanguage2025} to enhance the viewpoint robustness across diverse viewing conditions.  The other adopts active perception strategies---such as multi-view rotation---to proactively acquire more comprehensive environmental information and compensate for the inherent ambiguity of any single viewpoint. 
Open-vocabulary models are pre-trained predominantly on ground-level imagery and inherit limitations of ground views.  The performance in aerial views are unsatisfactory.  Active perception strategies incur time and energy costs, also lacking a principled criterion for evaluation.

\textbf{Research directions.}  For lack of aerial view in open-vocabulary models, aerial-specific datasets like the underexploited datasets identified in Section~\ref{sec:datasets} provide the visual diversity needed and fine-tune models. For active perception, the design of rational and standardized policies is required to optimize the information gain against the cost of delayed navigation. Additionally, high-fidelity 3D visual representations, like 3DGS or NeRF~\cite{NeRF2021}, render consistent features across arbitrary viewpoints, decoupling visual recognition from viewing geometry.

\subsection{Scalable Spatial Representation}
\label{sec:op_spatial}

\textbf{Problem.} The limited field of UAV view is contradiction with the long-range 3D spatial reasoning required for city-scale navigation (Section~\ref{sec:context})~\cite{xuGeoNavEmpoweringMLLMs2025}. Different spatial representations suit different scenarios. UAV navigation inherently involves a process of scene transition in Aerial VLN, which poses a fundamental challenge for spatial representation.

\textbf{Current approaches and limitations.} Three representation paradigms have been explored. Full 3D reconstruction provides the richest spatial information but is computationally expensive and difficult to maintain in real time~\cite{huang2026navdreamer}. 2D top-down projections (e.g., BEV maps~\cite{zhaoAerialVisionandlanguageNavigation2025}, semantic top-down maps~\cite{gaoAerialVisionandlanguageNavigation2024}) are efficient but discard crucial vertical dimension information. Abstract topological graphs~\cite{zhangCityNavAgentAerialVisionandlanguage2025, liuNavAgentMultiscaleUrban2024, xuGeoNavEmpoweringMLLMs2025} capture connectivity and high-level spatial structure but oversimplify and omit key landmark details needed for fine-grained navigation decisions.
No single representation resolves the trade-off between completeness and efficiency. All paradigms treat the representation as a static data structure, but real-world environments are dynamic.

\textbf{Research directions.}  The most promising path is lightweight hybrid spatial memory architectures that combine the efficiency of 2D/topological representations with selective 3D detail. For example, a system could maintain a coarse topological graph for global planning augmented with local 3D patches around the current position and upcoming landmarks, dynamically allocating representational detail based on navigational relevance. learning-based spatial memory approaches offer another avenue, though the ability to support precise metric reasoning remains to be demonstrated for aerial-scale environments.

\subsection{Continuous 6-DoF Action Execution}
\label{sec:op_action}

\textbf{Problem.} The UAV's action space is a 6-DoF continuous space (Section~\ref{sec:formulation}), yet the majority of current methods reduce this to a small set of discrete directional primitives (Section~\ref{sec:e2e}).  This simplification makes actions unnatural and imprecise, which creates a fundamental disconnect between the semantic intent expressed in the instruction and the motor commands required to realize that intent in continuous 3D space~\cite{liSkyVLNVisionandlanguageNavigation2025}.

\textbf{Current approaches and limitations.} Hierarchical methods (Section~\ref{sec:hierarchical}) address this partially by delegating continuous control to a low-level flight controller (NMPC~\cite{liSkyVLNVisionandlanguageNavigation2025}, Ego-Planner~\cite{huang2026navdreamer}, ViNT~\cite{shahViNTFoundationModel2023}) but introduce the planner-controller interface as a bottleneck (Section~\ref{sec:hierarchical}). VLA methods (Section~\ref{sec:e2e}) attempt end-to-end continuous control but remain in early stages, grappling with data scarcity.  Safety-aware methods like ASMA~\cite{sanyalASMAUnderlinetextAdaptiveUnderlinetextSafety2025} integrate control barrier functions to ensure dynamic feasibility but neglect some physical constraints of real UAV dynamics.

\textbf{Research directions.} First is differentiable planning-control pipelines, in which the high-level planner and the low-level controller are jointly optimized through a differentiable interface. Second is large-scale aerial demonstration data---particularly for VLA models with continuous actions---which can be potentially generated through expert policies in simulation (as in SINGER~\cite{adangSINGEROnboardGeneralist2025}) and then transferred to real platforms.  Third is integration of safety-aware control, ensuring that the output of the navigation policy is not merely semantically correct but physically realizable and safe.

\subsection{Onboard Deployment and Computational Efficiency}
\label{sec:op_onboard}

\textbf{Problem.}  The ultimate goal of Aerial VLN is complete and independent deployment on UAV platforms (Section~\ref{sec:context}).  Most existing methods rely on ground-station computation, controlling the UAV via communication links that introduce latency and single points of failure~\cite{zhangCityNavAgentAerialVisionandlanguage2025, sautenkovUAVCodeAgentsScalableUAV2025}.  The few works that attempt onboard inference report response frequencies too low for practical flight~\cite{wu2025vlaanefficientonboardvisionlanguageaction, chenGRADNAVVisionlanguageModel2026}.

\textbf{Current approaches and limitations.} Some LLM-centric Aerial VLN methods(Section~\ref{sec:e2e}) offloads heavy reasoning to a cloud model while running lightweight perception on the edge.  This reduces onboard computational demands but retains dependence on network connectivity.  Some hierarchical methods(Section~\ref{sec:hierarchical}) implicitly enable partial onboard deployment by running the low-level controller onboard while offloading the LLM planner to a ground station.
Current LLMs/VLMs require GPU resources that exceed what lightweight UAV platforms can carry.  Even quantized or distilled models struggle to meet the latency requirements of real-time flight. 

\textbf{Research directions.} Model distillation and pruning is essential, compact navigation-specialized models retain the reasoning capabilities of large models while fitting within onboard compute budgets. The synergy between a rational resource allocation framework and an efficient interface also contributes to latency reduction and sim-to-real onboard.

\subsection{Standardized Benchmarks and Reproducibility}
\label{sec:op_benchmarks}

\textbf{Problem.}  As presented in Sections~\ref{sec:datasets} and~\ref{sec:metrics}, current Aerial VLN benchmarks are fragmented with no unified simulation interface exists, datasets differ in action space definitions, success thresholds and evaluation protocols. This fragmentation makes rigorous cross-method comparison difficult.

\textbf{Current approaches and limitations.}  Several benchmarks have emerged independently, including AerialVLN~\cite{liuAerialVLNVisionandlanguageNavigation2023}, CityNav~\cite{leeCityNavLargescaleDataset2025}, OpenFly~\cite{gaoOpenFlyComprehensivePlatform2025a}, and OpenUAV~\cite{wangRealisticUAVVisionlanguage2024}, each serving specific task scenarios with their own conventions while also being adopted in some other Aerial VLN methods.  
But most benchmarks was still designed for its own research agenda, and no coordination mechanism exists to align definitions, metrics, or evaluation protocols across benchmarks. The result is that performance reported on different benchmarks is largely incommensurable (Section~\ref{sec:metrics}).

\textbf{Research directions.}  A unified Aerial VLN benchmark is needed that spans multiple environment types, action space configurations, and instruction complexities, along with standardized train/validation/test splits, evaluation metrics, and public leaderboards. Benchmark tasks require definition both in simulation and real-world environments, facilitating systematic measurement of the sim-to-real gap. Open-source baseline implementations that provide reproducible reference results on the unified benchmark also play a key role.

\subsection{From Single UAV to Swarm Intelligence}
\label{sec:op_swarm}

\textbf{Problem.}  Current Aerial VLN focuses exclusively on single UAV navigation. However, many of the most compelling application scenarios---large-area search and rescue, cooperative infrastructure inspection, distributed environmental monitoring---require coordinated navigation by multiple UAVs acting on shared or complementary instructions.  Scaling from single-UAV to multi-UAV Aerial VLN introduces challenges in decentralized decision-making, shared situational awareness, communication efficiency and collective task allocation.

\textbf{Current approaches and limitations.} Multi-agent methods (Section~\ref{sec:multiagent}) distribute reasoning across multiple LLM agents but operate a single physical UAV.  AeroDuo~\cite{wuAeroDuoAerialDuo2025} introduces a dual-UAV cooperative paradigm with high-altitude and low-altitude observation sharing, representing the closest existing work to multi-UAV Aerial VLN.  In UAV navigation, swarm coordination methods based on decentralized optimization~\cite{Xue2020ANS} and multi-agent RL exist but have not been integrated with VLN.
The multi-agent methods in Section~\ref{sec:multiagent} address cognitive distribution but not physical distribution.  Physical multi-UAV VLN introduces communication constraints, observation heterogeneity and coordination overhead.

\textbf{Research directions.} Advancing multi-UAV Aerial VLN requires aligning future efforts with its underlying internal technologies. First, decentralized planning represents a key direction that addresses the need for cooperative navigation through shared spatial representations, circumventing the bottlenecks of centralized control. Second, communication is another key direction, tackling the critical challenge of bandwidth explosion through multi-agent LLM architectures, avoiding the overhead of full reasoning.  Additionally, an effective language-guided task allocation module extends the instruction decomposition problem (Section~\ref{sec:op_longhorizon}) from a single-agent sequential plan to a multi-agent parallel plan, decomposing complex instructions into coordinated sub-tasks assigned to individual UAVs based on their perceptions.

\section{Conclusions}\label{sec:conclusions}


This survey provides a critical review of aerial vision-and-language navigation (Aerial VLN), from its formal foundations to the architectural diversity of current methods and the infrastructure that supports their development. Four principal findings emerge from the analysis.
 
First, the integration of LLMs and VLMs has demonstrably expanded the capability frontier of Aerial VLN.  Methods that leverage large pre-trained models, whether as end-to-end action predictors, high-level planners in hierarchical architectures, or reasoning agents in multi-agent systems, consistently handle longer instructions, more complex scenes, and more diverse environments than their task-specific predecessors.  However, this capability gain has not yet translated into reliable real-world performance: the best-reported success rates on challenging benchmarks remain well below what practical deployment demands, and nearly all results are confined to simulation.
 
Second, the discrete-versus-continuous action space divide remains the most consequential unresolved design choice in the field.  The majority of LLM-centric methods default to discrete directional primitives because they align naturally with token-based language model outputs, but this discretization is fundamentally at odds with the continuous 6-DoF dynamics of real UAV flight.  Hierarchical methods offer the most pragmatic current resolution by decoupling semantic planning from continuous control, and their compatibility with existing UAV autonomy stacks makes them the most deployment-ready architectural category.  End-to-end VLA methods represent the more ambitious long-term path but require substantially more demonstration data and training infrastructure than is currently available.
 
Third, the evaluation infrastructure is fragmented in ways that actively impede progress.  Inconsistent action space definitions, success thresholds, and evaluation protocols across benchmarks make cross-method comparison unreliable.  Critical dimensions of system performance, including inference latency, energy efficiency, safety violations, and instruction decomposition quality, are almost never measured.  The absence of a standardized simulation interface analogous to what Habitat provides for indoor VLN raises the barrier to entry and reduces reproducibility.  Addressing these infrastructure deficiencies is not merely a housekeeping task. It is a prerequisite for the field to move from demonstrating isolated capabilities to building cumulative, comparable knowledge.
 
Fourth, the gap between simulation results and real-world deployment remains the field's most formidable challenge.  Onboard computational constraints, communication latency, sensor noise, and dynamic environmental conditions are largely absent from current evaluations.  Closing this gap will require concurrent advances in lightweight model architectures, safety-aware control integration, and standardized sim-to-real benchmarks. None of these can be addressed by algorithmic innovation alone.
 
In summary, Aerial VLN stands at an inflection point.  The cognitive capabilities provided by LLMs have made it possible, for the first time, to build systems that can interpret complex natural language instructions and reason about 3D aerial environments at the city scale.  Whether these systems can be made efficient, robust, and safe enough to operate on physical UAV platforms, which transforms Aerial VLN from a simulation research topic into a deployed aerial intelligence capability, is the defining question for the next phase of the field.

\bibliographystyle{ieeetr}   
\bibliography{xxyreferences_elsevier}    

\end{document}